%% file: arxiv.tex
\documentclass{article}

\input{math_commands}

\usepackage[sort,authoryear,round]{natbib}
\usepackage{fullpage}
\usepackage[utf8]{inputenc} 
\usepackage[T1]{fontenc}    
\usepackage{xcolor}
\usepackage{hyperref}       
\definecolor{myblue}{rgb}{0,0.2,0.8}
\hypersetup{ %
pdftitle={},
pdfauthor={},
pdfsubject={},
pdfkeywords={},
pdfborder=0 0 0,
pdfpagemode=UseNone,
colorlinks=true,
linkcolor=myblue,
citecolor=myblue,
filecolor=myblue,
urlcolor=myblue,
pdfview=FitH}
\usepackage{authblk}
\usepackage{url}            
\usepackage{booktabs}       
\usepackage{amsfonts}       
\usepackage{amsthm}
\usepackage{nicefrac}       
\usepackage{microtype}      

\usepackage{graphicx}
\usepackage{xspace}
\usepackage[varqu,varl,var0,scaled=0.97]{inconsolata}
\usepackage{caption}
\usepackage{overpic}
\usepackage{wrapfig}
\usepackage{tcolorbox}
\usepackage{enumitem}
\usepackage{threeparttable}
\usepackage{subcaption}
\usepackage{array}
\usepackage{xcolor}
\usepackage{multirow}
\usepackage{multicol}
\usepackage{listings}
\usepackage{color}
\usepackage{lipsum}
\usepackage{bm}
\usepackage{amsmath} 
\usepackage{algorithm}
\usepackage{algpseudocode}
\usepackage{makecell}
\usepackage{float}
\usepackage{csquotes}

\definecolor{dkgreen}{rgb}{0,0.6,0}
\definecolor{gray}{rgb}{0.5,0.5,0.5}
\definecolor{mauve}{rgb}{0.58,0,0.82}

\newcommand{\name}{FocusDD\xspace}

\usepackage[capitalize,noabbrev]{cleveref}

\title{\bf{FocusDD: Real-World Scene Infusion\\ for Robust Dataset Distillation}}
\author[1]{Youbing Hu}
\author[2]{Yun Cheng}
\author[3,4]{Olga Saukh}
\author[2]{Firat Ozdemir}
\author[1]{Anqi Lu}
\author[1]{Zhiqiang Cao}
\author[1]{Zhijun Li}
\affil[1]{Faculty of Computing, Harbin Institute of Technology}
\affil[2]{Swiss Data Science Center, Zurich, Switzerland}
\affil[3]{Graz University of Technology, Austria}
\affil[4]{Complexity Science Hub Vienna, Austria}
\affil[ ]{\texttt{\{youbing, zhiqiang\_cao, luanqi\}@stu.hit.edu.cn, yun.cheng@sdsc.ethz.ch, saukh@tugraz.at, fozdemir@gmail.com, lizhijun\_os@hit.edu.cn}}

\begin{document}
\date{}
\maketitle

\begin{abstract}
Dataset distillation has emerged as a strategy to compress real-world datasets for efficient training. However, it struggles with large-scale and high-resolution datasets, limiting its practicality. 
This paper introduces a novel resolution-independent dataset distillation method \textbf{Focus}ed \textbf{D}ataset \textbf{D}istillation (\name), which achieves diversity and realism in distilled data by identifying key information patches, thereby ensuring the generalization capability of the distilled dataset across different network architectures. 
Specifically, FocusDD leverages a pre-trained Vision Transformer (ViT) to extract key image patches, which are then synthesized into a single distilled image. These distilled images, which capture multiple targets, are suitable not only for classification tasks but also for dense tasks such as object detection. To further improve the generalization of the distilled dataset, each synthesized image is augmented with a downsampled view of the original image. Experimental results on the ImageNet-1K dataset demonstrate that, with 100 images per class (IPC), ResNet50 and MobileNet-v2 achieve validation accuracies of 71.0\% and 62.6\%, respectively, outperforming state-of-the-art methods by 2.8\% and 4.7\%. Notably, FocusDD is the first method to use distilled datasets for object detection tasks. On the COCO2017 dataset, with an IPC of 50, YOLOv11n and YOLOv11s achieve 24.4\% and 32.1\% mAP, respectively, further validating the effectiveness of our approach. 
\end{abstract}

\section{Introduction}
\label{sec:intro}

Contemporary deep learning has achieved remarkable success largely due to the exponential growth in model sizes \citep{he2016deep,szegedy2015going,dosovitskiy2020image,radford2021learning} and data scales \citep{deng2009imagenet,ridnik2021imagenet,kirillov2023segment}. This growth has led to the development of advanced neural networks that achieve groundbreaking performance in tasks like image classification \citep{dosovitskiy2020image}, object detection \citep{carion2020end}, and natural language processing \citep{vaswani2017attention}. 
However, this progress is not without its challenges. The rapid expansion of model complexities and data volumes has led to significantly increased computational costs and time expenses, in particular when training large neural networks on high-resolution and large-scale datasets \citep{liu2021swin,touvron2021training,NEURIPS2021_9a49a25d}. 
These challenges significantly hinder the practical deployment of deep learning models, especially in resource-limited environments \citep{ignatov2019ai}.

\begin{wrapfigure}{r}{9.0cm} 
    \begin{center}
        \centering
    \includegraphics[width=0.5\textwidth]{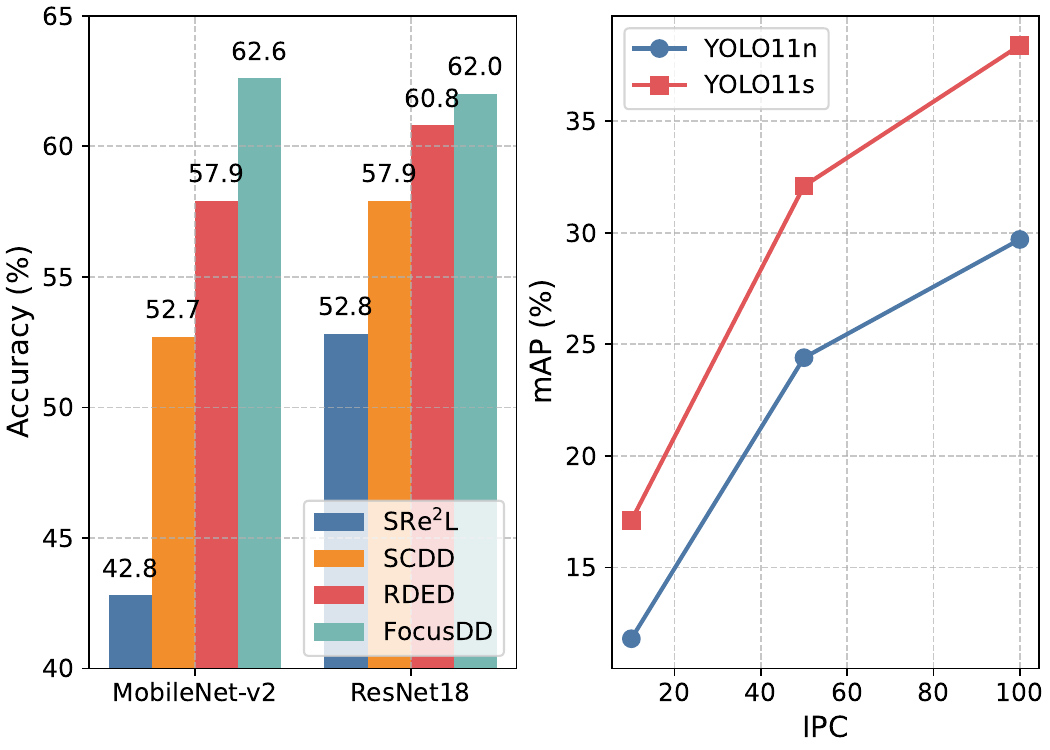} 
    \captionof{figure}{FocusDD performance on classification and detection tasks. \textbf{Left:} For classification with IPC=100, we use MobileNet-v2~\citep{sandler2018mobilenetv2} and ResNet-18~\citep{he2016deep} as validation models to evaluate the ImageNet-1K~\citep{deng2009imagenet} validation set. SCDD~\citep{SCDD}, SRe$^{2}$L~\citep{yin2024squeeze}, and RDED~\citep{sun2024diversity} are the current SOTA methods. \textbf{Right:} In the detection task, we use YOLOv11~\citep{khanam2024yolov11} as the validation model to evaluate the COCO2017~\citep{lin2014microsoft} validation set. FocusDD is the first method to explore dataset distillation for object detection tasks.}
    \label{fig:1}
    \end{center}
\end{wrapfigure}

  

  \begin{table}[ht]
  \centering
  \begin{tabular}{c | c | c | c | c}
      \hline
      Model & Method & Flower102& Food101 & CIFAR100 \\
      \hline
      \multirow{3}{*}{ResNet18} & Random & 22.4 & 57.8 & 54.5 \\
      & RDED & 67.8 & 74.2 & 69.3 \\
      & FocusDD & \textbf{71.1} & \textbf{77.6} & \textbf{71.3} \\
      \hline
  \end{tabular}
  \caption{We evaluate the generalization performance of ResNet-18~\citep{he2016deep} as a validation model trained on  distilled data. With IPC set to 10, the model is first pre-trained on a dataset distilled by RDED~\citep{sun2024diversity} and FocusDD, then fine-tuned on the original data for 10 epochs. The datasets used are Flowers-102~\citep{nilsback2008automated}, Food-101~\citep{bossard2014food}, and CIFAR-100~\citep{krizhevsky2009learning}. ``Random'' refers to a model trained directly on the target datasets for 10 epochs without pre-training.}
  \label{tab:1}
\end{table}

\begin{figure*}[h]
  \centering
  \includegraphics[width=1.0\textwidth,height=0.5\textwidth]{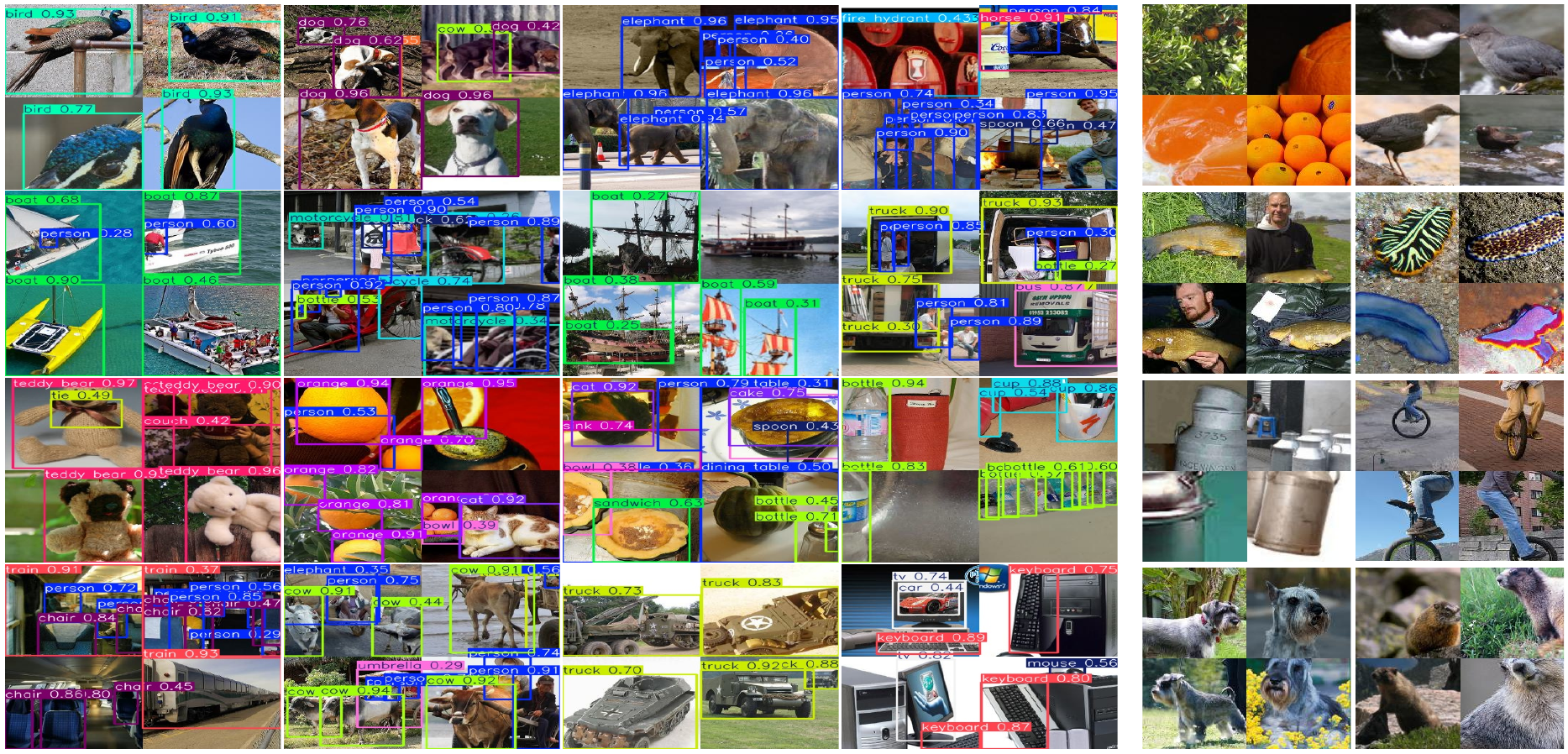}
  \caption{Visualization of the FocusDD-distilled images on different tasks.
\textbf{Left}: Visualization of training samples for object detection using FocusDD-distilled images. Using YOLOv11x~\citep{khanam2024yolov11} as the teacher model, soft supervision is applied to train YOLOv11n and YOLOv11s, tested on the COCO2017~\citep{lin2014microsoft} validation set. The numbers in each image correspond to COCO categories. \textbf{Right}:  Visualization of training samples for classification using FocusDD-distilled images. Soft supervision with ResNet-18~\citep{he2016deep} as the teacher guides ResNet-18 and MobileNet-v2~\citep{sandler2018mobilenetv2} training, tested on the ImageNet-1K~\citep{deng2009imagenet} validation set. The performance is shown in Fig.~\ref{fig:1}.}
  \label{fig:motivation}
\end{figure*}

Dataset distillation \citep{wang2018dataset} has emerged as a promising strategy to address these challenges. The core idea is to compress large, real-world datasets into smaller, more manageable representations that retain essential information while reducing the computational burden of ingesting them. 
Various methods have been proposed, including coreset selection-based distillation \cite{toneva2018empirical, feldman2020neural, paul2021deep, meding2021trivial, tan2024data}, which select representative samples from the original dataset; bi-level optimization-based distillation \citep{zhang2023accelerating,zhao2023dataset,du2023minimizing,guo2023towards}, which treats dataset distillation as a meta-learning problem involving two nested optimization loops—where the outer loop optimizes the meta-dataset and the inner loop trains a model with the distilled data; 
and distillation with prior regularization \citep{cazenavette2023generalizing,lu2023can,cui2023scaling}, which leverages prior knowledge at the feature level to guide the generation of the condensed dataset.

Although traditional solutions have made significant progress in handling small-scale and low-resolution datasets (such as Tiny-ImageNet~\citep{le2015tiny}, downscaled ImageNet~\citep{chrabaszcz2017downsampled}, or subsets of ImageNet~\citep{kim2022dataset}), their high computational cost limits their practical application when scaled to high-resolution and large-scale datasets. To address this issue, SRe$^{2}$L~\citep{yin2024squeeze} proposed a decoupled approach for model updates and datasets, which was the first to extend dataset distillation techniques to the scale of ImageNet. Subsequently, several methods \citep{SCDD,sun2024diversity,yin2023dataset,loolarge} have been proposed to improve the efficiency of SRe$^{2}$L and significantly enhance accuracy. For example, SCDD~\citep{SCDD} replaces the batch-level statistics used in SRe$^{2}$L with statistics calculated over the entire distillation dataset. RDED~\citep{sun2024diversity} randomly crops a region from the original high-resolution image, selects multiple images with the highest authenticity scores, and merges them into a distilled image. While these methods effectively synthesize high-resolution images, they rely on specific network architectures during the distillation process, limiting the generalization ability of the distilled dataset. Furthermore, the datasets distilled by these methods typically only apply to classification tasks and cannot be directly applied to dense tasks, such as object detection.

In this paper, we propose a novel dataset distillation method called Focused Dataset Distillation (FocusDD), which aims to improve the efficiency and realism of dataset distillation by focusing on key information patches within the data. FocusDD consists of two stages: (i) information extraction and (ii) image reconstruction.
In the information extraction stage, we leverage a pre-trained Vision Transformer (ViT) \citep{dosovitskiy2020image} to guide the selection of key image patches. By using ViT, we can accurately extract key image patches corresponding to foreground objects, thereby enhancing the realism of the distilled dataset and ensuring the relevance of the extracted information. Since these distilled images contain target regions, they are well-suited for downstream dense tasks such as object detection. As shown in Fig.~\ref{fig:1}, FocusDD demonstrates superior performance at different IPC levels on the COCO validation dataset when using the YOLOv11 model; Fig.~\ref{fig:motivation} visualizes the training samples of FocusDD distillation images across different tasks. This is the first work to extend dataset distillation methods to object detection tasks.
In the reconstruction stage, we combine downsampled versions of representative real images with the extracted key image patches to generate distilled images. This process not only preserves the diversity of the dataset but also ensures its realism, providing high-quality training data that enhances the generalization ability of the model. Table~\ref{tab:1} highlights the advantages of FocusDD in improving model generalization performance. Finally, an optional dynamic fine-tuning on a small subset of the original dataset can further boost performance and is investigated in Appendix~\ref{appendix:dft_1}.

Overall, this paper makes the following contributions to the field of dataset distillation:

\begin{itemize}
    \item We are the first to integrate ViTs into the image distillation process. By selectively emphasizing critical regions and foreground objects, ViT ensures that the distilled dataset retains crucial contents of the data distribution for effective model training.
    \item  Our method not only preserves the realism and diversity of the images but also enables effective application to downstream dense tasks, such as object detection. By leveraging Attention-guided distillation, we can clearly identify the image regions most critical for model learning. To the best of our knowledge, we are the first work to extend dataset distillation to object detection tasks.
    \item  We provide a rigorous evaluation of our approach including multiple ablation studies and show improved model generalization capabilities across different network architectures. Compared to SOTA methods on classification tasks, FocusDD improves the accuracy of ResNet50 and MobileNetV2 at IPC level 50 by 2.8\% and 4.7\%, respectively. On object detection tasks, FocusDD achieves 24.4\% mAP with YOLOv11n, and 32.1\% mAP with YOLOv11s at an IPC of 50 on the COCO validation set.
\end{itemize}

\section{Related Work}
\label{related_work}

Data distillation~\citep{wang2018dataset} aims to reduce the computational costs of training deep learning models by condensing large datasets into smaller, information-rich subsets. Most previous dataset distillation methods~\citep{zhou2022dataset,wang2018dataset,nguyen2021dataset,cazenavette2022dataset,zhao2023dataset,wang2022cafe,lee2022dataset,zhao2020dataset,guo2023towards} focus on small-scale and low-resolution datasets~\citep{le2015tiny,chrabaszcz2017downsampled,kim2022dataset} and can be classified into several categories: 
Bi-level optimization methods treat dataset distillation as a meta-learning problem, where an outer loop optimizes the synthetic dataset while an inner loop focuses on model training using distilled data, methods include FRePo~\citep{zhou2022dataset}, DD~\citep{wang2018dataset}, RFAD~\citep{nguyen2021dataset}, KIP~\citep{nguyen2021dataset}, and LinBa~\citep{deng2022remember}. 
Trajectory-matching methods align model training trajectories on the original and distilled datasets over multiple iterations, methods include MTT~\citep{cazenavette2022dataset}, TESLA~\citep{cui2023scaling}, and DATM~\citep{guo2023towards}.
Distribution-matching methods match the distribution of the distilled dataset with that of the original in a single optimization step, with examples like KFS~\citep{lee2022dataset}, DM~\citep{zhao2023dataset}, CAFE~\citep{wang2022cafe}, HaBa~\citep{liu2022dataset}, and IT-GAN~\citep{zhao2022synthesizing}. 
Gradient-matching methods align gradients of the network trained on original and synthesized data, with examples including DSA~\citep{zhao2021dataset}, IDC~\citep{kim2022dataset}, DC~\citep{zhao2020dataset}, and DCC~\citep{lee2022dataset}.

Building on these foundations, recent approaches have extended dataset distillation to large-scale, high-resolution datasets. For example, SRe$^{2}$L~\citep{yin2024squeeze} decouples model updates and dataset synthesis through ``squeeze", ``restore". and ``relabel" stages, pioneering the expansion of dataset distillation to ImageNet-scale resolutions. SCDD~\citep{SCDD} further improves on SRe$^{2}$L by replacing batch-level statistics with global dataset statistics, achieving notable performance gains. D3S~\citep{loolarge} reframes dataset distillation as a domain shift problem, introducing a scalable algorithm, while RDED~\citep{sun2024diversity} generates distilled images by randomly cropping and selecting high-realism image regions. Additionally, some dataset distillation methods~\citep{gu2023efficient,su2024generative} employ the concept of diffusion models for distilling datasets.


Although previous methods excel with high-resolution images, they compress the original dataset into a specific architecture~\citep{SCDD,yin2024squeeze,sun2024diversity}, limiting the generalization of the distilled dataset. In contrast, FocusDD synthesizes datasets using the well-established Attention mechanism, which improves generalization, as shown in Table~\ref{tab:1} and Table~\ref{tab:generalization} across different ViT models. Furthermore, by synthesizing images focused on target locations, FocusDD extends its use to dense tasks like object detection, marking the first application of dataset distillation in this domain.

\section{Approach}
\label{sec:approach}

We first provide background knowledge on dataset distillation and ViT in Sec.~\ref{sec:Preliminaries}. Next, we give a detailed description of our method \name in Sec.~\ref{sec:Focused_dataset_dis}, along with a theoretical analysis in Appendix~\ref{theory}. Finally, we discuss how to train models using the distilled dataset in Sec.~\ref{sec:model_training_on_distilled}.

\begin{figure*}
  \centering
  \includegraphics[width=\textwidth,height=0.28\textwidth]{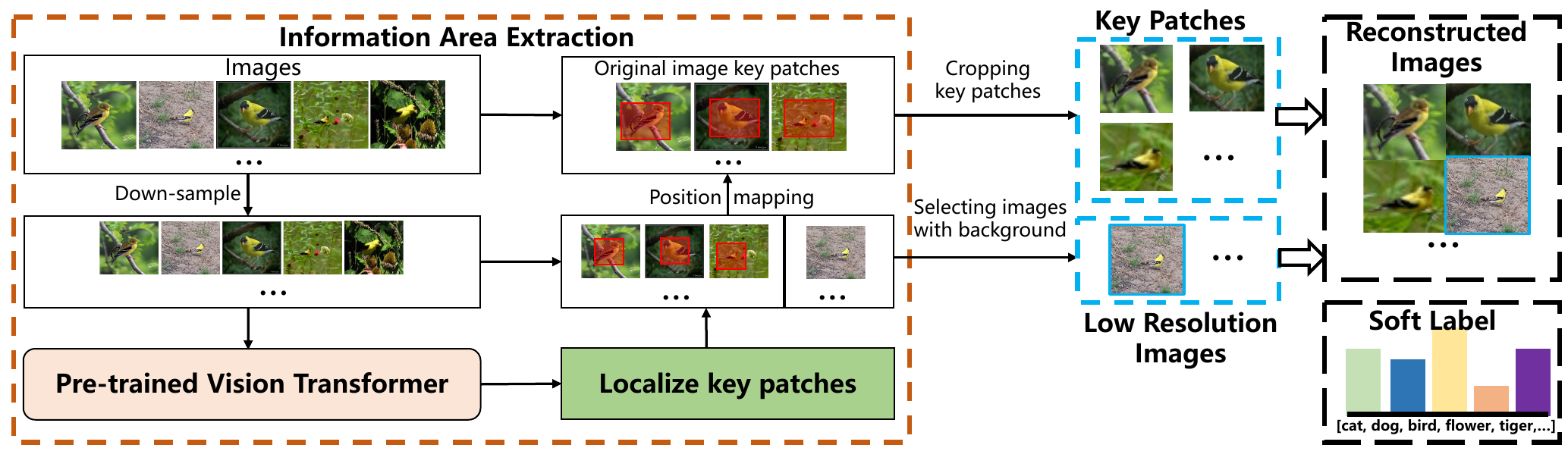}
  \caption{Overview of the \name framework. \name comprises two main stages: information extraction and image reconstruction. In the information extraction stage, a pre-trained ViT model guides the selection of key patches, identifying those containing key patches and representative real images with background details. During the image reconstruction stage, these patches are combined with images rich in background information to reconstruct a compiled, realistic image. Subsequently, these images are relabelled using a model with the same architecture as the validation model.}
  \label{fig:system_framwork}
\end{figure*}

\subsection{Preliminaries}
\label{sec:Preliminaries}

\textbf{Data Distillation/Condensation.} 
Dataset distillation~\citep{wang2018dataset} aims to compress information from a large-scale original dataset to a new compact dataset while striving to preserve the utmost degree of the original data informational essence.
The resulting compressed dataset denoted as \( D' \), should enable a model trained on it to perform comparably to a model trained on the original, full dataset \( D \).
Considering a large labeled dataset \( D = \{(\bm{x}_1, y_1), \dots, (\bm{x}_{\vert D \vert}, y_{\vert D \vert})\} \), where \( \vert D \vert \) denotes the total number of samples, and each \( \bm{x}_i \) 
is an image with its corresponding label \( y_i \). The aim is to create a condensed dataset \( D' = \{(\bm{\tilde{x}}_1, \tilde{y}_1), \dots, (\bm{\tilde{x}}_{\vert D^\prime \vert}, \tilde{y}_{\vert D^\prime \vert})\} \) that retains the key features of \( D \), with \( {\vert D^\prime \vert} \ll {\vert D \vert} \), ensuring that this reduction in size does not compromise the dataset integrity. The learning objective is to minimize the performance disparity between the model trained on \( D' \) and the one trained on \( D \), as expressed by the following constraint:
\begin{equation}
     \mathrm{sup}\big\{\big|\ell(\phi_{\bm{\theta}_{D}}(\bm{x}), y) - \ell(\phi_{\bm{\theta}_{D^\prime}}(\bm{x}), y)\big|\big\}_{(\bm{x}, y)\sim D}\leq \epsilon,
\end{equation}
where \( \epsilon \) represents the allowable performance disparity between models trained on \( D' \) versus those trained on \( D \). Here, \( \bm{\theta}_D \) parameterizes the neural network \( \phi \), optimized on \( D \) as follows:
\begin{equation}
    \bm{\theta}_D = \mathop{\arg\min}\limits_{\bm{\theta}}\mathbb{E}_{(\bm{x}, y)\in D}[\ell(\phi_{\bm{\theta}}(\bm{x}), y)].
\end{equation}
In this formulation, \( \ell \) is the loss function, and \( \bm{\theta}_{D'} \) is defined in a similar manner for the condensed dataset. This framework ensures that \( D' \) maintains the essential characteristics of \( D \), allowing effective training on a smaller scale.

\textbf{Vision Transformer.} 
Vision Transformer (ViT) \citep{dosovitskiy2020image} adapts the Transformer architecture \citep{vaswani2017attention}, originally developed for natural language processing, to the domain of image analysis. They treat image patches as sequential inputs, allowing the model to capture global dependencies across the image. Each image is segmented into patches, which are embedded and supplemented with positional encodings to maintain spatial information, denoted as: \( \bm{x} = [\bm{x}_{\text{cls}}; \bm{E}(\bm{p}_1); \bm{E}(\bm{p}_2); \ldots; \bm{E}(\bm{p}_K)] + \bm{E}_{\text{pos}} \),
where \( \bm{E} \) is the embedding function, \( \bm{p}_i \) are the patches, \( \bm{x}_{\text{cls}} \) is the class token, and \( \bm{E}_{\text{pos}} \) represents the positional encodings.
The self-attention mechanism then calculates attention scores to determine the relevance of each patch relative to others:
\begin{equation}
\begin{aligned}
    \mathcal{A}(\mathbf{Q}, \mathbf{K}) = \text{Softmax}(\frac{\mathbf{QK}^T}{\sqrt{d}}) = [\mathbf{A}^1; \mathbf{A}^2; \dots; \mathbf{A}^K],  \\
    \text{Attention}(\mathbf{Q}, \mathbf{K}, \mathbf{V}) = \mathcal{A}(\mathbf{Q}, \mathbf{K})\mathbf{V},
\end{aligned}
\end{equation}
where \( \mathbf{Q} \), \( \mathbf{K} \), and \( \mathbf{V} \) are the query, key, and value matrices from \( \bm{x} \), \( d \) is the embedding dimension of $\mathbf{K}$, and \( K \) is the number of patches. 
The average attention score \( \bm{s} \) for an image reflects the outcome of a single-head self-attention mechanism. In multi-head self-attention, scores from all attention heads are averaged to yield the final image attention score. The class token \( \bm{x}_{\text{cls}} \) is processed by a classifier \( \mathcal{F} \) to derive the category prediction distribution \(\textbf{p}^c\):
\begin{equation}
\begin{aligned}
   \bm{s} = \frac{1}{K}\sum_{k=1}^K \mathbf{A}^k = [s^1, s^2, \dots, s^K], \\
   \textbf{p}^c = \mathcal{F}(\bm{x}_{\text{cls}}) = [p_1^c, p_2^c, \dots, p_C^c],
\end{aligned}
\end{equation}
where \( C \) indicates the number of categories.

\subsection{Focused Dataset Distillation with Attention}
\label{sec:Focused_dataset_dis}
This section introduces \name, a dataset distillation method that reconstructs compiled images by focusing on the target and representative background information of real images. Fig.~\ref{fig:system_framwork} and Algorithm~\ref{alg:1} in Appendix~\ref{appendix:algorithm} provide an overview. Further details are provided below.

\begin{figure*}
  \centering
  \includegraphics[width=\textwidth,height=0.2\textwidth]{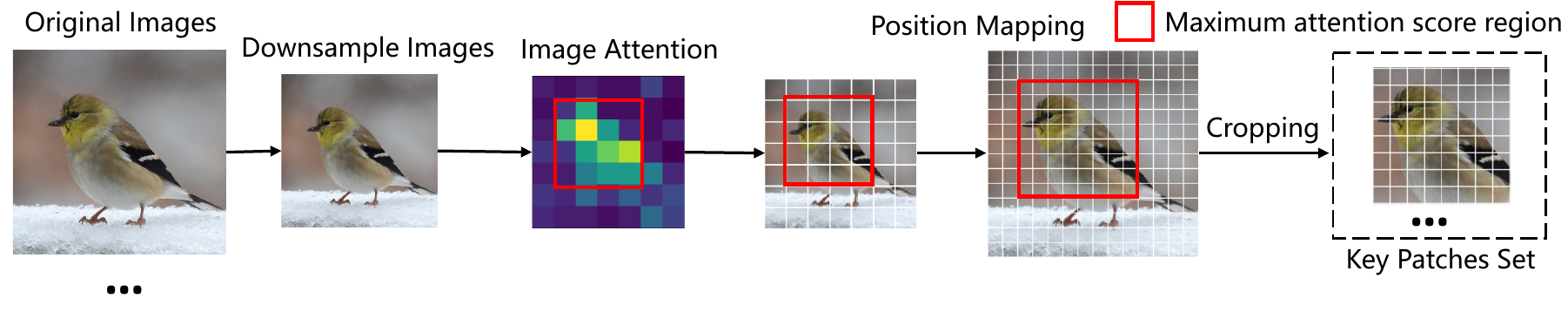}
\caption{The FocusDD process of selecting key image patches. Downsampling greatly reduces the computational cost of dataset distillation (see Table~\ref{tab:com_cost} in the Appendix~\ref{appendix:dft_2}) and allows the direct use of downsampled images to improve the generalization performance of the synthesized dataset (see Table~\ref{fig:1}).}

  \label{fig:regional_selection}
\end{figure*}

\textbf{Attention-guided Information Extraction.}
We utilize an attention mechanism to identify and extract regions with the highest attention scores from multiple images, thereby compiling images with enhanced detail. These regions are then combined to form a detailed composite image set, as illustrated in Fig.~\ref{fig:system_framwork}.
The process initiates by performing the following steps on each image \(\bm{x}_i \in \mathbb{R}^{H \times W \times Ch}\) within each category-specific subset \(D_c\) of the dataset \(D\): each \(\bm{x}_i\) is downsampled to \(\bm{x}'_i\) and segmented into non-overlapping patches of size \(P \times P\). This downsampling produces \(K = \frac{H}{P} \times \frac{W}{P}\) patches per image, which are subsequently reorganized into the structured form \(\mathbb{R}^{\frac{H}{P} \times \frac{W}{P} \times P^2 Ch}\), with each row and column representing a token.
These tokens are embedded and fed into a pre-trained ViT model, yielding predictive distributions \(\mathbf{p}_i^c\) and attention scores \(\bm{s}_i \in \mathbb{R}^K\). Likewise, we reorganize each attention score \(\bm{s}_i\) into the format \(\frac{H}{P} \times \frac{W}{P}\).
To determine the size of the highest attention score region for each image \(\bm{x}'_i\), we introduce an adjustable hyperparameter \(\alpha\), which specifies the number of patches \(\lfloor \alpha\frac{H}{P} \times \alpha\frac{W}{P} \rfloor\). We then introduce a realism score \(s_i^{\text{real}}\) to identify the key patch for each image. Specifically, our realism score combines the prediction distribution \(\textbf{p}_i^c\) of each image with the highest attention region score \(s_i^{\text{area}}\), defined as follows:
\begin{equation}
\label{real_scores}
    s_i^{\text{real}} = \text{max}(\text{softmax} (\mathbf{p}_i^c)) + \eta s_i^{\text{area}},
\end{equation}
where \(\eta\) is a balancing factor. 
Intuitively, $s_i^{\text{real}}$ indicates the need to select a representative image with a focus on the target region within it. This implies that our selection process should prioritize images that represent the overall scene accurately and emphasize the specific area of interest, ensuring that the target region is well-captured and highlighted in the chosen image.

After calculating the realism score \(s_i^{\text{real}}\), we associate each score with its corresponding image \(D_c\) and sort the scores in descending order. 
Based on these scores, we select the top-$M$ images from the sorted \(D_c\) and extract the regions with the highest attention scores.
The center indices of these high attention regions are determined using the following formula:
\begin{equation}
\label{eq:7}
    (i, j) = \mathop{\arg\max}_{i, j}\sum_{p,q}\bm{s}^{i + p - \lfloor \frac{h}{2} \rfloor, j + q - \lfloor \frac{w}{2} \rfloor},
\end{equation}
where \(h = \lfloor \alpha\frac{H}{P} \rfloor\), \(w = \lfloor \alpha\frac{W}{P} \rfloor\), \(p \in \{0, 1, \ldots, h-1\}\), and \(q \in \{0, 1, \ldots, w-1\}\).
Utilizing the positional mapping function \(\rho\), we translate these indices to the dimensions of the original image \(\bm{x}_i\), marking the key information region \(\bm{x}_i^\star\) in the high-resolution image as:
\begin{equation}
\begin{aligned}
\label{eq:8}
    \bm{x}_i^\star = \text{area}(\rho((i - \lfloor \alpha\frac{H}{2P} \rfloor, j + \lfloor \alpha\frac{W}{2P} \rfloor), \\ (i + \lfloor \alpha\frac{H}{2P} \rfloor, j - \lfloor \alpha\frac{W}{2P} \rfloor ))).
\end{aligned}
\end{equation}
Finally, we compile the identified key patches into a set \(\tilde{T}_c = \{\bm{x}_i^\star\}_{i=1}^M\), where each sample \(\bm{x}_i^\star\) is a crop of the high-resolution image containing fine details, thereby preserving maximum informational content for use in the compiled composites. 
To further enhance the diversity of the synthesized images, we randomly select \( N \) low-resolution sampled images from \( D_c \) that were not chosen as key information patches. These images are weighted based on their prediction confidence scores and added to \( T'_c = \{\bm{x}_i^\prime\}_{i=1}^N \) as a background information set.
Fig.~\ref{fig:regional_selection} illustrates the process of selecting the set of key patches.

\textbf{Information Reconstruction.}
The size of key patches is typically smaller than the target distilled images. Directly using these key patches as distilled images can result in sparse information distribution in the pixel space, thereby reducing the effectiveness of the learning model \citep{yun2021re,shen2022fast,yin2024squeeze}. As shown in Table~\ref{tab:patch_number_effect}, using distilled image sets composed solely of key patches leads to a decreased model performance. Therefore, we combine the set of images containing key information patches \(\tilde{T}_c\) with the set of low-resolution images \(T_c^\prime\) to supplement the class category \(c\) information with the typical context in which they appear.
Specifically, 
we randomly select \(m\) patches from \(\tilde{T}_c\) and \(n\) low-resolution images from \(T_c^\prime\) each time. The selected images are then concatenated to compile the final composite image \(\tilde{\bm{x}_j}\):
\begin{equation}
\label{eq:9}
    \bm{\tilde{x}}_j = \text{concat} (( \{\bm{x}_j^\star\}_{j=1}^{m} \subset \tilde{T}_c),
    (\{\bm{x}_j^\prime\}_{j=1}^{n} \subset T_c^\prime )).
\end{equation}

By default, we set the combined total of patches and images to $m + n = 4$ (see Fig.~\ref{fig:patch_num} in the Appendix~\ref{appendix:dft_2}), where $m = 3$ represents the selection of three patches from the key information patch collection $\tilde{T}_c$, and $n = 1$ corresponds to selecting one low-resolution image from the background information collection $T_c^\prime$ (Table~\ref{tab:patch_number_effect}).
Following the RDED \citep{sun2024diversity} and SRe$^2$L \citep{yin2024squeeze}, we apply a soft label approach \citep{shen2022fast} to the compiled images. This method generates region-level soft labels \( \tilde{y}_j^k = \ell(\phi_{\bm{\theta}_{D^\prime}}(\bm{\tilde{x}}_j^{k})) \), where \( \bm{\tilde{x}}_j^{ k} \) is the $k$-th region in the distilled image, and \(\tilde{y}_j^{k}\) is its corresponding soft label. 

By iterating over each category \(c\) in \(D\), performing the information extraction and image reconstruction processes, and adding the generated images and labels \(\{ \bm{\tilde{x}}_j,y_j\}\) to the distilled dataset \(D^\prime\), we ultimately obtain the complete distilled dataset \(D^\prime \). 



\subsection{Model Training on Distilled Datasets}
\label{sec:model_training_on_distilled}
After assembling the distillation dataset $D^\prime$, we initiate training of a student model \( \phi_{\bm{\theta}_S} \) from random initialization using this dataset, in line with strategies proposed by \citet{yin2024squeeze} and \citet{sun2024diversity}. For classification tasks, the training employs a cross-entropy loss function defined as:
\begin{equation}
    \mathcal{L} = -\sum_j\sum_k \tilde{y}^k_j\log\phi_{\bm{\theta}_S}(\bm{\tilde{x}}^k_j).
\end{equation}

To optimize training efficiency for the detection task, we input the distilled images into YOLOv11x~\citep{khanam2024yolov11} to compute the classification and bounding box losses and supervise model updates using Kullback–Leibler divergence loss. To accelerate training, we use YOLOv11x to generate ground truth (GT) boxes for each synthesized image and train the model following the standard YOLOv11 procedure.

In Appendix~\ref{appendix:dft_1}, we outline how a model, initially trained on a distilled dataset, undergoes Dynamic Fine-Tuning (DFT) on the data obtained by dynamically sampling the original dataset. This method leads to further performance enhancements across all architectures.

\section{Experiments}
\label{sec:experiments}

\subsection{Experimental Setup}

\textbf{Datasets and Implementation Details.} We conducted rigorous and extensive validation of FocusDD on the large-scale ImageNet-1K dataset \citep{deng2009imagenet} to comprehensively evaluate its performance. The ImageNet-1K dataset consists of approximately 1.2 million training images with a resolution of 224$\times$224 pixels, spanning 1000 categories. For key patch extraction, we utilized the Deit-S model \citep{touvron2021training}, pre-trained by \citet{hu2024lf}.
We maintain a constant side ratio \(\alpha\) of 0.8 and \(\eta\) of 30.
We set the value of \( N \) equal to IPC and \( M \) equal to 3$\times$IPC, effectively limiting the size of the distillation dataset to the total number of pixels in the IPC image.
We train target models including ResNet-$\{18, 50, 101\}$ \citep{he2016deep}, MobileNet-v2 \citep{sandler2018mobilenetv2}, and EfficientNet-b0 \citep{tan2019efficientnet} to validate the distilled datasets. All models are trained on the distilled dataset for 300 epochs with 224$\times$224 image resolution.
Our experiments were conducted using an NVIDIA 4090 GPU.
Additional experimental details and Tiny-ImageNet \citep{le2015tiny} experiments are provided in Appendix~\ref{implementation_details} and Table~\ref{tab:tiny_imagenet_res} Appendix~\ref{appendix:dft_1}, respectively.

\begin{table*}\tiny
  \caption{Comparison with SOTA baseline dataset distillation methods on the ImageNet-1K dataset. Following the revalidation model, we present the accuracy (\%) achieved by various architectures on the full ImageNet-1K dataset. Our method significantly outperforms all compared baseline methods. The table highlights the \textbf{highest accuracy in bold} and \underline{underlines the second-highest accuracy}. For the SCDD~\citep{SCDD}, D3S~\citep{loolarge}, and GVBSM~\citep{shao2023generalized} methods, we list the results reported in the original papers.
} 
  \label{tab:imagenet_res}
  \centering
  \small{
  \begin{tabular}{c|cccc | cccc}
    \toprule

         \multirow{2}{*}{  Method}   &  \multicolumn{8}{c}{  IPC }   \\
       \cmidrule(r){2-9}
        & 1 & 10 & 50 & 100 & 1 & 10 & 50 & 100 \\
    \midrule
    & \multicolumn{4}{c}{  ResNet-18 (69.8 $\pm$0.1)  } & \multicolumn{4}{c}{  ResNet-50 (76.2$\pm$0.1) } \\
     \cmidrule(r){2-9}
     SRe$^{2}$L & 0.1$\pm$0.1 & 21.3$\pm$0.6& 46.8$\pm$0.2 & 52.8$\pm$0.3  & 0.3$\pm$0.1 & 28.4$\pm$0.1& 55.6$\pm$0.3 & 61.0$\pm$0.4\\
     SCDD & - & 32.1$\pm$0.2 & 53.1$\pm$0.1 & 57.9$\pm$0.1 & - & 38.9$\pm$0.1 & 60.9$\pm$0.2 & 65.8$\pm$0.1 \\
     GVBSM & - & 31.4$\pm$0.5 & 51.8$\pm$0.4 & 55.7$\pm$0.4 & - & 35.4$\pm$0.8 & 58.7$\pm$0.3 & 62.2$\pm$0.3\\
     RDED & \underline{6.6$\pm$0.2} & \underline{42.0$\pm$0.1} & 56.5$\pm$0.1 & 60.8$\pm$0.4 & 
     \underline{5.7$\pm$0.1} & \underline{42.3$\pm$0.3} & 64.8$\pm$0.6 & \underline{68.2$\pm$0.2}\\
     D3S & - & 39.1$\pm$0.3 & \underline{60.2$\pm$0.1} & \textbf{63.0$\pm$0.2} & - & 41.9$\pm$0.7 & \underline{65.8$\pm$0.1} & \underline{68.2$\pm$0.1} \\
    
    \name  & \textbf{8.8$\pm$0.2} & \textbf{45.3$\pm$0.1} & \textbf{61.7$\pm$0.1} & \underline{62.0$\pm$0.2} & \textbf{6.8$\pm$0.1} & \textbf{46.3$\pm$0.2} & \textbf{69.1$\pm$0.3} & \textbf{71.0$\pm$0.1} \\

    \midrule
    & \multicolumn{4}{c}{  MobileNet-V2 (71.8$\pm$0.1)  } & \multicolumn{4}{c}{   EfficientNet-B0 (76.3$\pm$0.1) } \\
    \cmidrule(r){2-9}
     SRe$^{2}$L & 0.3$\pm$0.1 & 10.2$\pm$2.6 & 31.8$\pm$0.3 & 42.8$\pm$0.6  & 0.4$\pm$0.2 & 11.4$\pm$2.5 & 34.8$\pm$0.4&  49.6$\pm$0.5   \\
     RDED & \underline{4.9$\pm$0.6} & \underline{33.8$\pm$0.6} &\underline{54.2$\pm$0.2}&\underline{57.9$\pm$0.6}  & \underline{3.4$\pm$0.2} & \underline{33.3$\pm$0.9} & \underline{57.7$\pm$0.1} &\underline{63.7$\pm$0.3}\\
     \name & \textbf{5.1$\pm$0.1} & \textbf{34.6$\pm$0.1} & \textbf{58.7$\pm$0.3} & \textbf{62.6$\pm$0.1 } & \textbf{4.8$\pm$0.2} & \textbf{40.1$\pm$0.2} & \textbf{60.7$\pm$0.1} & \textbf{66.6$\pm$0.3} \\

    \bottomrule
  \end{tabular}
  }
\end{table*}

\textbf{Evaluation and Baselines.} 
We compare our approach with several SOTA methods for distilling large-scale, high-resolution datasets, including SRe$^2$L~\citep{yin2024squeeze}, SCDD~\citep{SCDD}, GVBSM~\citep{shao2023generalized}, D3S~\citep{loolarge} and RDED~\citep{sun2024diversity}. 
In our evaluation process, we generate a unique distillation dataset for each IPC level (1, 10, 50, 100) for \name and reuse it across multiple network architectures.

\subsection{Performance Evaluation}
\label{Comparison_with_SOTA_Approaches}

\textbf{ImageNet-1K Classification.} 
Tables~\ref{tab:imagenet_res} and~\ref{tab:imagenet_res_r101} present the experimental results of FocusDD on the ImageNet-1K dataset, showing its significant advantages across various architectures (e.g., ResNet-18, ResNet-50, ResNet-101, MobileNet-V2, EfficientNet-B0) and IPC settings. FocusDD consistently outperforms other methods, especially for low IPCs (1, 10, and 50), achieving higher accuracy, which is crucial for scenarios with limited samples or resource constraints. For instance, on ResNet-18, FocusDD achieves accuracies of 8.8\% and 45.3\% at IPCs of 1 and 10, respectively, significantly surpassing RDED and D3S.
Even for higher IPCs (e.g., IPC = 100), FocusDD maintains strong performance, often achieving or nearing the best results on ResNet-50 and EfficientNet-B0. This demonstrates FocusDD's ability to excel under minimal and small-sample data conditions, adapting effectively across different models and IPC configurations. 

\begin{table}[!t]\small
  \caption{Accuracy comparison (\%) of SOTA baseline dataset distillation methods using ResNet101 (77.4$\pm$0.2) on ImageNet-1K.
  } 
  \label{tab:imagenet_res_r101}
  \centering
  \begin{tabular}{c|cccc }

    \toprule
    \multirow{2}{*}{  Method}   &  \multicolumn{4}{c}{  IPC }   \\
       \cmidrule(r){2-5}
        & 1 & 10 & 50 & 100 \\
\midrule
  SRe$^{2}$L~\citep{yin2024squeeze} & 0.6$\pm$0.1 & 30.9$\pm$0.1 & 60.8$\pm$0.5 &62.8$\pm$0.2  \\

     SCDD~\citep{SCDD} & - & 39.6$\pm$0.4 & 61.0$\pm$0.3 & 65.6$\pm$0.2 \\
     GVBSM~\citep{shao2023generalized} & - & 38.2$\pm$0.4 & 61.0$\pm$0.4 & 63.7$\pm$0.2 \\
     RDED~\citep{sun2024diversity} & \underline{5.9$\pm$0.4} & \underline{42.1$\pm$1.0} & 61.2$\pm$0.4 & \underline{69.5$\pm$0.5}\\
    D3S~\citep{loolarge} & - & 42.1$\pm$3.8 & \underline{65.3$\pm$0.5} & 68.9$\pm$0.1 \\
     \name & \textbf{8.5$\pm$0.2} &\textbf{43.1$\pm$0.2}& \textbf{69.9$\pm$0.2}& \textbf{72.9$\pm$0.1} \\

    \bottomrule
  \end{tabular}
\end{table}

\begin{table}[!t]\small
\caption{Comparison of classification accuracy (\%) when training with diffusion-based network generated datasets and FocusDD. ResNet-18 was used as a validation model.}
\label{tab:diffusion_methods}
\centering
\begin{tabular}{c|ccc}

\toprule
IPC & DiT \citep{peebles2023scalable} & MinmaxDiffusion \citep{gu2023efficient} & \textbf{\name} \\
\midrule
10 & 39.6$\pm$0.4  &44.3$\pm$0.5 & \textbf{45.3$\pm$0.1} \\
50 & 52.9$\pm$0.6 &58.6$\pm$0.3 & \textbf{61.7$\pm$0.1} \\

\bottomrule
\end{tabular}
\end{table}

Additionally, we compare our method with diffusion-based image generation models \citep{peebles2023scalable, gu2023efficient} in Table~\ref{tab:diffusion_methods}. 
Appendix~\ref{appendix:coreset_selection} compares FocusDD with Coreset-based selection methods \citep{welling2009herding, forgy1965cluster} on ImageNet-1K, showing consistent superiority of FocusDD. Table~\ref{tab:tiny_imagenet_res} in Appendix~\ref{appendix:dft_1} shows FocusDD’s strong performance on Tiny-ImageNet, even at low IPCs, aligning with results on ImageNet-1K.


\noindent\textbf{COCO Object Detection.} In the object detection task, we use YOLOv11x~\citep{khanam2024yolov11} as the teacher model to perform soft-supervised training on YOLOv11n and YOLOv11s models from scratch for a total of 100 epochs, with all experimental settings following the official YOLOv11~\citep{khanam2024yolov11} configuration. Fig.~\ref{fig:1} shows the mAP performance of the FocusDD-distilled dataset on the COCO validation set under different IPC settings. The figure indicates that as IPC increases, model performance also gradually improves. For example, when IPC is 50, YOLOv11s achieves an mAP of 32.1\%. FocusDD performs effectively on object detection tasks because distilled images are composed of multiple patches containing targets, each of which may include objects of interest to the detection model.

\subsection{Performance  Analysis}

\begin{table}[!t]\small 
\caption{Impact of different ViT models on FocusDD accuracy.  }
\label{tab:generalization}
\centering
\begin{tabular}{c|cccc}
\toprule
\multirow{2}{*}{ \shortstack{Distillation \\ Architecture }  }&    \multicolumn{4}{c}{  IPC } \\
\cmidrule(r){2-5}
 & 1& 10 & 50 & 100 \\
\midrule
Deit-S & 8.8$\pm$0.2  &45.3$\pm$0.1 & 61.7$\pm$0.1 & 62.0$\pm$0.2\\
LV-ViT-S  & 9.4$\pm$0.3 & 45.8$\pm$0.2 & 62.3$\pm$0.2 & 62.8$\pm$0.1 \\

\bottomrule
\end{tabular}
\end{table}

\begin{figure}[!t]
    \begin{center}
        \includegraphics[width=0.5\linewidth]{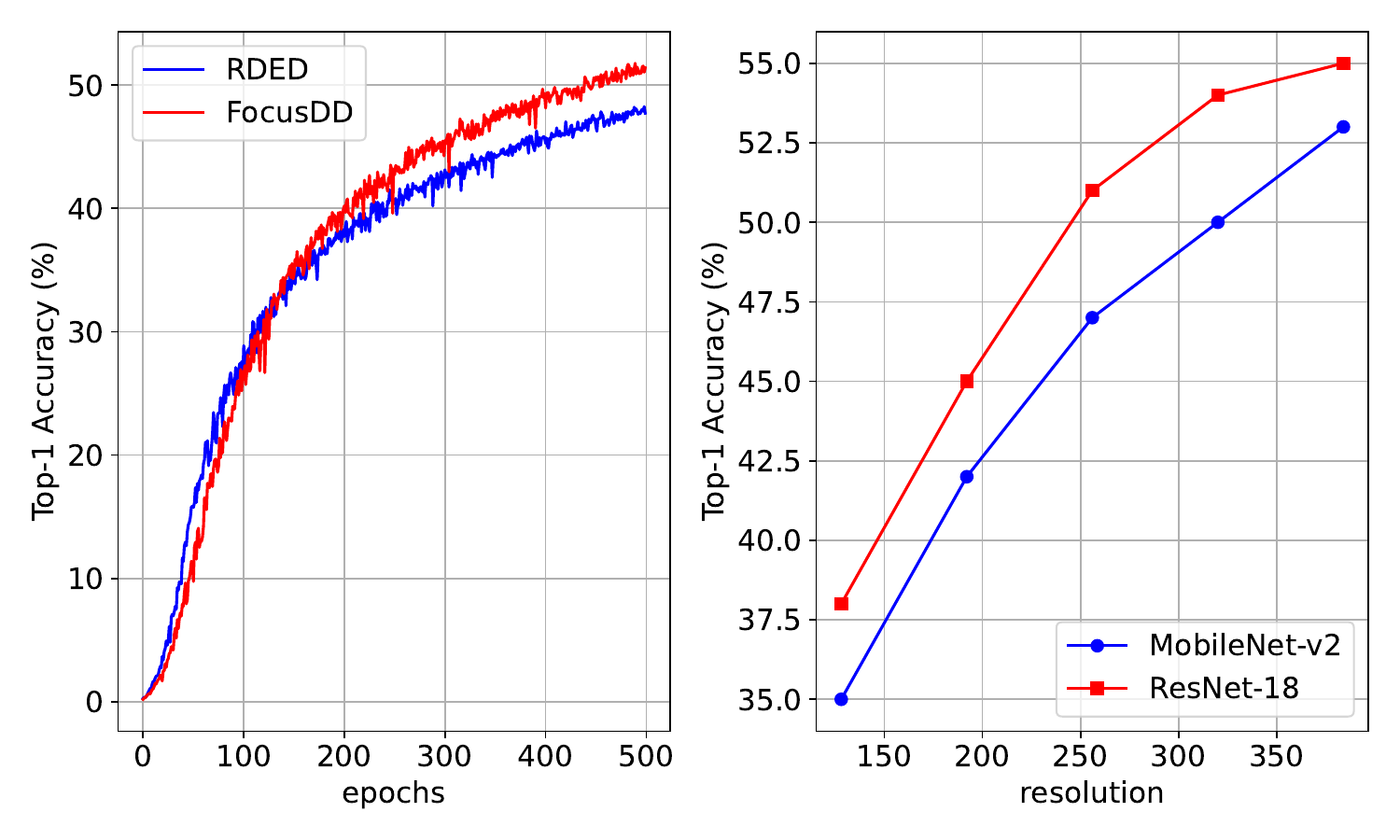}
        \caption{Model accuracy with varying epoch and resolution. \textbf{Left:} Accuracy changes with training epochs using ResNet-18 as the validation model in the IPC-10 setting. \textbf{Right:} The impact of the image resolution of synthetic dataset on model accuracy.}
        \label{fig:adaptive_resolution_accuracy}
    \end{center}
\end{figure}

\begin{table*}[t]\small 
  \caption{Effectiveness of different technologies in our method on ImageNet-1K. ResNet-18 is used as the validation model with an IPC of 10. From left to right, each column represents an incremental addition of technologies starting with the base method: Coreset Filtering (CF), Add Background Information (ABI), Extracting Key Patches (EKP), Image Reconstruction (IR), and Labels Reconstruction (LR).}
  \label{tab:Each_Technique_Effectiveness}
  \centering
  \begin{tabular}{c|ccccc}
    \toprule
    FocusDD (Base)     & +CF & +ABI & +EKP & +IR & +LR  \\
    \midrule
    Accuracy (\%) & 18.4$\pm$0.3 & 23.6$\pm$0.1 & 28.2$\pm$0.2 & 30.9$\pm$0.1 & 45.3$\pm$0.1  \\
    \bottomrule
  \end{tabular}
\end{table*}

\begin{table*}[t]\small 
  \caption{Comparing key patch selection strategies using various metrics, including Herding \citep{welling2009herding}, K-Means \citep{forgy1965cluster}, and Realism \citep{sun2024diversity}, which are current SOTA methods. 
  All methods are evaluated using ResNet-18 on ImageNet-1K with IPC=10.}
  \label{tab:patch_selection_strategies}
  \centering
  \begin{tabular}{c|ccccccc}
    \toprule
     Method       & Random & Herding & K-Means & Realism & Min-AS & R-AS & Max-AS\\
    \midrule
   Accuracy (\%)   & 37.9$\pm$0.5  & 38.4$\pm$0.1  & 38.2$\pm$0.1 & 42.0$\pm$0.1 & 41.6$\pm$0.3 & 42.6$\pm$0.8 & \textbf{45.3$\pm$0.1} \\

    \bottomrule
  \end{tabular}
\end{table*}

\begin{table*}[!t]\small 
\caption{The effect of the number of patches in each compiled image. Each synthesized image includes 4 patches, with $m$ key patches and $n$ low-resolution background images. We used ImageNet-1K and MobileNet-v2 with IPC=10 to evaluate different patch configurations. }
\label{tab:patch_number_effect}
\centering
\begin{tabular}{c|cccccc}

\toprule
Patches & $m=4, n=0$ & \bm{$m=3, n=1$} & $m=2, n=2$& $m=1, n=3$ & $m=0, n=4$ \\
\midrule
Accuracy & 32.6$\pm$0.3 & \textbf{34.6$\pm$0.1} & 34.2$\pm$0.2 & 33.2$\pm$0.2 & 31.8$\pm$0.5 \\

\bottomrule
\end{tabular}
\end{table*}

\textbf{Cross-Architecture Generalization.}  
Table~\ref{tab:generalization} evaluates the impact of different ViT models on FocusDD's performance on ImageNet-1K, using ResNet-18 for validation. 
The results demonstrate that our method maintains consistent performance across ViT architectures,
corroborating the idea that the attention-based key patch selection in FocusDD is similarly effective for also different transformer architectures.
Table~\ref{fig:1} presents results from fine-tuning pre-trained models for 10 epochs on CIFAR-100~\citep{krizhevsky2009learning}, Flowers-102~\citep{nilsback2008automated}, and Food-101~\citep{bossard2014food}, with datasets distilled using different methods. Our method shows superior generalization and enhances downstream task performance.


\noindent\textbf{Training Epoch and Adaptive Resolution Synthesis.}
There is a lack of a unified benchmark for comparing methods, such as the number of training steps or used image resolution. This makes it hard to compare all SOTA baselines in the form they were initially presented to the community; e.g., those in \citep{gu2023efficient} and \citep{kim2022dataset}, conducted across 1000 epochs. Nonetheless, we show the impact of training time on FocusDD and RDED in Fig.~\ref{fig:adaptive_resolution_accuracy} (left). Accuracy improves with more training epochs, consistent with the findings of D3S~\cite{loolarge}. Fig.~\ref{fig:adaptive_resolution_accuracy} (right) shows the FocusDD synthesis accuracy at different resolutions, with accuracy improving as resolution increases, demonstrating the effectiveness of adaptive resolution control in image synthesis. Additionally, Table~\ref{tab:higher_resolution} in Appendix~\ref{appendix:dft_2} illustrates the impact of input ViT resolution on the curated dataset. Notably, during training, all images are resized to a fixed resolution of 224$\times$224.

\noindent\textbf{Qualitative Analysis.} Fig.~\ref{fig:imagenet_visualization} in Appendix~\ref{Visualization_Synthetic_Data} visualizes compiled images generated by different SOTA methods. SRe$^2$L, SCDD, and GVBSM produce blurrier images, likely due to overreliance on specific models during dataset compression, which hampers generalization. In contrast, RDED and our \name method generate more realistic images by cropping key patches from real image locations. Unlike RDED, our method includes both key patches and contextual backgrounds, enhancing realism and diversity. The attention mechanism used in our method, validated in the vision community \citep{hu2024lf,chen2023smmix,chen2023cf,rao2021dynamicvit}, improves interpretability and offers deeper insights into dataset distillation. 


\subsection{Ablation study}

\textbf{Effectiveness of Each Technique in \name.} To validate the effectiveness of all components within our \name, we conduct ablation studies for each of them.
Table~\ref{tab:Each_Technique_Effectiveness} illustrates that all techniques employed in \name are essential for
achieving a remarkable final performance. We observed that label reconstruction at the patch level significantly improves accuracy, consistent with the findings of previous methods \citep{sun2024diversity,SCDD,yin2024squeeze}. 


\noindent\textbf{Effectiveness of Selecting Key Patches Through Realism Score.}
Table~\ref{tab:patch_selection_strategies} demonstrates the effectiveness of different key patch selection strategies using realism scores. Our method, which utilizes the maximum attention score (Max-AS) as a score metric, surpasses all compared methods. Specifically, Max-AS achieves a 14.3\% accuracy improvement over the current SOTA methods—Herding \citep{welling2009herding}, K-Means \citep{forgy1965cluster}, and Realism \citep{sun2024diversity}. Compared to its variants, the minimum attention score (Min-AS) and random attention score (R-AS), Max-AS achieves the highest accuracy by focusing on target regions while selecting the same key patches and representative low-resolution images.

\noindent\textbf{Impact of the number of patches in compiled images.}
By adjusting the number of key patches \( m \) and the number of low-resolution images \( n \), each compiled image is composed of \( m \) key patches and \( n \) low-resolution images. 
We adopt the combination that achieves the highest accuracy as our default setting, namely, composing the final image with three key patches and one low-resolution image containing global information. Table~\ref{tab:blancing_factor} in Appendix~\ref{appendix:dft_2} shows the impact of the balancing factor $\eta$ on \name's performance. We select $\eta = 30$ as the default value.


\section{Conclusion}
In this paper, we introduce \name, a novel method that employs attention mechanisms to guide data distillation effectively for large-scale and high-resolution datasets. \name extracts key patches from image target regions, ensuring critical information and realism, and combines them with low-resolution contextual backgrounds to create distilled images for training. This diversifies the dataset and enhances model generalization. Additionally, \name is invariant to the resolution of target images, making it a flexible and performant choice for data distillation regardless of the underlying image resolution requirements. Extensive experiments and ablation studies demonstrate \name's effectiveness and offer insights into applying deep learning to large-scale data and complex models for both classification and object detection tasks.

\clearpage
\bibliography{arxiv.bib}

\newpage
\appendix
\section*{Appendix}
\input{arxiv_appendix}

\end{document}

%% file: math_commands.tex
\usepackage{amsmath,amsfonts,bm}

\def\eqref#1{equation~\ref{#1}}

\def\1{\bm{1}}

\DeclareMathAlphabet{\mathsfit}{\encodingdefault}{\sfdefault}{m}{sl}
\SetMathAlphabet{\mathsfit}{bold}{\encodingdefault}{\sfdefault}{bx}{n}



%% file: arxiv_appendix.tex


In the appendix, we provide details omitted in the main text, including:
\begin{itemize}[leftmargin=*]
 \item Section \ref{implementation_details}: Implementation Details.
 \item Section \ref{appendix:algorithm}: Focused Dataset Distillation Algorithm.
 \item Section \ref{appendix:further_experiments}: Further Experimental Results.
 \item Section \ref{theory}: Theoretical Analysis.
 \item Section \ref{Visualization_Synthetic_Data}: Sample Visualizations of Synthetic Data.
\end{itemize}

\section{Implementation Details}
\label{implementation_details}

\subsection{Pre-training ViT Models}
For the ImageNet-1K dataset, we directly use the model pre-trained by LF-ViT \citep{hu2024lf}, which is based on the implementations of Deit-S \citep{hu2024lf} and LV-ViT-S \citep{NEURIPS2021_9a49a25d}. This model performs inference at both the standard resolution of 224$\times$224 and a higher resolution of 288$\times$288, efficiently extracting crucial information patches for dataset distillation. To further reduce inference time, we disable the Focus stage in the LF-ViT implementation. More details and features of LF-ViT can be found on the official website.
For the lower resolution and smaller scale Tiny-ImageNet dataset, we train a modified version of the Deit-S-based LF-ViT \citep{hu2024lf} from scratch to extract key information patches. Specifically, we reduce the model's depth to 4 layers, set the patch size to 4$\times$4, adjust the embedding dimension to 192, and reduce the number of heads to 3. This modified model is trained from scratch using the same hyperparameters as those used for ImageNet-1K.

\subsection{\name Implementation Details}
We maintain a fixed side ratio \(\alpha = 0.8\) and a balancing factor \(\eta = 30\) for both the ImageNet-1K and Tiny-ImageNet datasets. To compile each image \(\bm{\tilde{x}}_j\) in the distilled dataset \(D^\prime\), we set \(N\) and \(M\) to IPC and 3$\times$IPC, respectively. The compile process involves concatenating three key patches from the key information collection \(\tilde{T}_c\) and one low-resolution background 
image from \(T_c^\prime\), resulting in the compiled image as described by Eq.~\ref{eq:9}. For instance, at an IPC of 100, we select 300 key information patches and 100 downsampled low-resolution images with background information, ensuring the synthesis of a diverse and representative image. This approach adapts to different IPC values to accurately reflect the dataset's variability.
Aligned with techniques from SRe$^2$L \citep{yin2024squeeze} and RDED \citep{sun2024diversity}, we employ Fast Knowledge Distillation \citep{shen2022fast} to relabel distilled images. Each distilled image \(\tilde{\bm{x}}_j\) is randomly cropped into several patches, with their coordinates recorded within \(\tilde{\bm{x}}_j\). Soft labels \(\tilde{y}_j^k\) are generated and stored for each $k$-th patch. These labels are aggregated to construct a comprehensive label \(\tilde{y}_j\) for each image, facilitating nuanced and accurate labeling reflective of the diverse visual features captured in the compiled images.

\begin{table}
\centering
\makeatletter\def\@captype{table}
\begin{tabular}{l|l}

    \toprule
    Config     & Value   \\
    \midrule
    Optimizer & SGD     \\
    Base learning rate     & 0.2      \\
    Weight decay     & 1e-4 \\
    Optimizer momentum & 0.9 \\
    Batch size &  256 \\
    Learning rate schedule &  Cosine decay \\
    Training epoch &  300 \\
    Augmentation & RandomResizedCrop \\
    \bottomrule
\end{tabular}
\caption{Tiny-ImageNet training hyper-parameters.}
\label{tab:tiny_hyper_setting}
\end{table}

\begin{table}

\makeatletter\def\@captype{table}
\centering
\begin{tabular}{l|l}

    \toprule
    Config     & Value   \\
    \midrule
    Optimizer & AdamW     \\
    Base learning rate     & 0.001      \\
    Weight decay     & 0.01 \\
    Optimizer momentum & $\beta_1, \beta_2 = 0.9, 0.999$ \\
    Batch size &  128 \\
    Learning rate schedule &  Cosine decay \\
    Training epoch &  300 \\
    Augmentation & RandomResizedCrop \\
    \bottomrule
\end{tabular}
\caption{ImageNet-1K training hyper-parameters.}
\label{tab:image1k_hyper_setting}
\end{table}

\textbf{Training on Distilled Dataset.} 
We use a model with the same architecture as the validation model, pre-trained on the corresponding original and full datasets, to generate soft labels for the synthesized images. For Tiny-ImageNet, our teacher model is pre-trained on the complete Tiny-ImageNet dataset, following the hyperparameters in \citep{yin2024squeeze}. When training the validation model on the distilled Tiny-ImageNet dataset, we use the hyperparameters shown in Table~\ref{tab:tiny_hyper_setting}. For ImageNet-1K, all teacher models use pre-trained models from the \texttt{torchvision} library. When training the validation model on the distilled ImageNet-1K dataset, we follow the parameters in Table~\ref{tab:image1k_hyper_setting}. Both datasets are augmented by CutMix with a mix probability \(p = 1.0\) and a beta distribution \(\beta = 1.0\).

For the object detection task, we selected samples from the ImageNet-1K dataset corresponding to the categories in COCO2017~\citep{lin2014microsoft} and generated a dataset based on the IPC settings. YOLOv11x~\citep{khanam2024yolov11} was used as the teacher model to annotate this dataset. Then, YOLOv11s and YOLOv11n were trained from scratch on the annotated dataset, and their performance was evaluated on the COCO2017 validation set. All training hyperparameters were kept identical to the official YOLOv11~\citep{khanam2024yolov11} configuration.

\textbf{Dynamic Fine-Tuning Parameter Settings.}
During the Dynamic Fine-Tuning (DFT) process (detailed in Appendix~\ref{appendix:dft_1}), we randomly select images with the same IPC from the original dataset in each iteration to form a new dataset for fine-tuning. The hyperparameters for fine-tuning match those used for training the validation model on the synthesized dataset. We set the learning rate to 0.00025, with 50 epochs and a batch size of 64. The learning rate for MobileNet-v2 during DFT is set to 0.001.

\section{Focused Dataset Distillation Algorithm}
\label{appendix:algorithm}
Algorithm~\ref{alg:1} outlines \name's key patch information extraction and image reconstruction. In the implementation, these tasks are executed in batches, allowing for parallel processing. Table~\ref{tab:synthesis_time_and_memory_consumption} shows the time required to synthesize 100 images.

\section{Further Experimental Results}
\label{appendix:further_experiments}

\subsection{Coreset Selection}
\label{appendix:coreset_selection}

\begin{table}\small
\centering
\caption{\textbf{Comparison of different Coreset selection-based
dataset distillation baselines.} All methods use ResNet-18 as the validation model and IPC=10. }
\label{tab:coreset}
\begin{tabular}{c|c|c|c|c}

\toprule
Dataset & Random & Herding & K-Means & \textbf{\name} \\
\midrule
Tiny-ImageNet & 7.5$\pm$0.1  &9.0$\pm$0.3 & 8.9$\pm$0.2 & \textbf{51.5$\pm$0.1}\\
ImageNet-1K& 4.4$\pm$0.1 &5.8$\pm$0.1 & 5.5$\pm$0.1 & \textbf{45.3$\pm$0.1} \\

\bottomrule
\end{tabular}
\end{table}

\textbf{Comparison with Coreset Selection Baselines.}  
In this evaluation, we use ResNet-18 as a validation model on the ImageNet-1K dataset with IPC set to 10, comparing it to a dataset extraction strategy based on coreset selection.
We evaluate the top-1 validation accuracy achieved by three distinct Coreset selection methodologies: (1) Random selection; (2) Herding, as introduced by \citet{welling2009herding}; and (3) K-Means clustering, based on \cite{forgy1965cluster}. The results detailed in Table~\ref{tab:coreset} demonstrate that the performance of these selection techniques is significantly compromised when applied independently for dataset distillation. In contrast, our \name achieves substantial accuracy improvements of 39.5\% on ImageNet-1K and 38.5\% on Tiny-ImageNet, respectively.

\begin{algorithm*}
\caption{Focused Dataset Distillation with Attention}
\label{alg:1}
\textbf{Input:} Dataset $D$, pre-trained ViT model, $\alpha$, $\eta$, $M$, $N$, $m$, $n$ \\
\textbf{Output:} Distilled dataset $D'$

\begin{algorithmic}[1]
\For{each category-specific subset $D_c \subset D$}
    \For{each image $\bm{x}_i \in D_c$}
        \State Downsample $\bm{x}_i$ to $\bm{x}'_i$ and segment into non-overlapping patches of size $P \times P$
        \State Embed patches and feed into ViT model
        \State Obtain predictive distributions $\mathbf{p}_i^c$ and attention scores $\bm{s}_i \in \mathbb{R}^K$
        \State Use the predefined $\alpha$ to determine the size of the patch
        \State Calculate realism score $s_i^{\text{real}}$ by Eq.~\ref{real_scores} and associate it with the corresponding image $\bm{x}_i$.
    \EndFor
    \State Sort image $D_c$ by each image's realism score in descending order
    \State Select the top-$M$ images and obtain the center indices of the key patch regions by Eq.~\ref{eq:7}
    \State Extract key patches $\bm{x}_i^\star$ by Eq.~\ref{eq:8}
    \State Add key patches into set $\tilde{T}_c$

    \State Randomly select $N$ downsampled low-resolution images in $D_c$ from non-top-$M$ images
    \State Add selected downsampled low-resolution images to set $T'_c$

    \For{$\bm{x}_m^\star \in \tilde{T}_c$ and $\bm{x}_n^\prime \in T_c^\prime$ }
        \State Randomly select $m$ key patches from $\tilde{T}_c$ and $n$ downsampled images from $T_c^\prime$
        \State Concatenate to compile composite image $\bm{\tilde{x}}_j$ by Eq.~\ref{eq:9}
        \State Apply soft label approach to $\bm{\tilde{x}}_j$
        \State Add $\{ \bm{\tilde{x}}_j, y_j \}$ to distilled dataset $D'$
    \EndFor
\EndFor

\State \textbf{return} Distilled dataset $D'$
\end{algorithmic}
\end{algorithm*}

\subsection{Dynamic Fine-Tuning} 
\label{appendix:dft_1}

Following the training of model \(\phi_{\bm{\theta}_S}\) on the distilled dataset \(D^\prime\), we implement the Dynamic Fine-Tuning (DFT) process. The DFT process involves fine-tuning the model on subsets of the original dataset that are dynamically sampled at each epoch. To preserve consistency with the structural properties of the synthetic dataset, images are randomly selected at an IPC level from each category to form new datasets for fine-tuning. This strategy is systematically applied throughout each epoch, introducing variability and generating a unique dataset for fine-tuning in every cycle. This approach significantly enhances the diversity of the data without additional training overhead, thereby boosting the model's generalization ability across diverse data representations. Furthermore, the DFT methodology not only capitalizes on the attributes of synthetic data but also closely aligns the model's performance with real-world data distributions, culminating in notable enhancements in performance.

\textbf{ImageNet-1K Datsset.} 
Table~\ref{tab:imagenet_res_appendix}  presents the experimental results of training FocusDD for 1000 epochs and combining it with DFT on the ImageNet-1K dataset.
We find that DFT further improves the performance of \name across all architectures. In particular, when IPC=100, \name + DFT demonstrates exceptionally small declines in accuracy—0.7\%, 1.9\%, 1.0\%, 2.8\%, and 1.8\% across the evaluated models—almost achieving performance equivalent to training with the complete dataset. These minimal accuracy losses highlight the robustness of \name when augmented by DFT, effectively leveraging the combined strengths of focused data distillation and iterative fine-tuning. The success of this approach underscores that merging \name with DFT offers a powerful and efficient strategy for minimizing accuracy losses in high-scale learning environments, making it particularly suitable for scenarios where resources are limited but high performance is imperative.

\textbf{Tiny-ImageNet Dataset.} 
Table~\ref{tab:tiny_imagenet_res} evaluates our method, \name, integrated with DFT on the Tiny-ImageNet dataset, showing similar trends as observed with the ImageNet-1K dataset. Notably, using EfficientNet-b0 at an IPC of 100, \name not only matches but also exceeds the performance of baseline models by 1.1±0.1\%. This improvement likely stems from DFT's random selection of IPC samples each round, enhancing the diversity of training data and thus boosting performance. This result highlights the benefits of combining \name with DFT to optimize performance under data constraints.

\subsection{Additional Experiments}
\label{appendix:dft_2}
\begin{table*}\small
  \caption{Our \name incorporates dynamic fine-tuning to further improve performance. It is worth noting that to further highlight the accuracy improvement brought by dynamic fine-tuning, the accuracy of FocusDD is based on the results after training for 1000 epochs.} 
  \label{tab:imagenet_res_appendix}
  \centering
  \begin{tabular}{cc|cccc}
    \toprule

       \multirow{2}{*}{  Architecture}&  \multirow{2}{*}{  Method}   &  \multicolumn{4}{c}{  IPC }   \\
       \cmidrule(r){3-6}
       & & 1 & 10 & 50 & 100 \\
    \midrule
    \multirow{2}{*}{  ResNet-18 (69.8)} 
    
    & \name & {10.7$\pm$0.2} & {52.5$\pm$0.1} & {63.1$\pm$0.1} & {68.0$\pm$0.2} \\
     & \name + DFT & {14.7$\pm$0.1} & {57.6$\pm$0.1} & {65.8$\pm$0.2} & {69.1$\pm$0.1}\\

     \midrule
    \multirow{2}{*}{  ResNet-50 (76.2)} 
    & \name  & {6.92$\pm$0.1} & {56.5$\pm$0.2} & {70.1$\pm$0.3} & {71.1$\pm$0.2}\\
    & \name + DFT  & {12.3$\pm$0.2} & {62.9$\pm$0.2}& {72.8$\pm$0.2} & {74.3$\pm$0.2}  \\

     \midrule
    \multirow{2}{*}{  ResNet-101 (77.4)} 

    & \name & {7.3$\pm$0.2} &{53.8$\pm$0.2}& {71.5$\pm$0.2}& {73.5$\pm$0.1} \\
    & \name + DFT & {14.7$\pm$0.3} & {58.3$\pm$0.2} & {72.6$\pm$0.3} & {76.4$\pm$0.1}  \\

    \midrule
    \multirow{2}{*}{  MobileNet-V2 (71.8)} 
    & \name &{8.4$\pm$0.1} & {49.5$\pm$0.1} & {61.6$\pm$0.3} & {66.0$\pm$0.1 } \\
    & \name + DFT & {12.1$\pm$0.3} & {56.0$\pm$0.1} & {66.4$\pm$0.1} & {69.0$\pm$0.2}  \\

    \midrule
    \multirow{2}{*}{  EfficientNet-B0 (76.3)} 
    & \name & {12.7$\pm$0.2} & {50.4$\pm$0.2} & {67.9$\pm$0.1} & {68.5$\pm$0.2} \\
   & \name +DFT & {17.6$\pm$0.4} &  {59.9$\pm$0.2} & {73.4$\pm$0.1} &  {74.5$\pm$0.2} \\
    \bottomrule
  \end{tabular}
\end{table*}

\begin{table*}[!t]\small 
\caption{Impact of $\eta$ on FocusDD performance. We use MobileNet-v2 as the validation model on the ImageNet-1k dataset, with IPC set to 10.}
\label{tab:blancing_factor}
\centering
\begin{tabular}{c|cccccccc}

\toprule
$\eta$ & 0 & 10 & 20 & \textbf{30} & 40 & 50 & 100 \\
\midrule
Accuracy & 32.4$\pm$0.2 & 33.1$\pm$0.2 & 34.2$\pm$0.2 & \textbf{34.6$\pm$0.1} & 34.2$\pm$0.2 & 33.6$\pm$0.4 & 32.8$\pm$0.3 \\

\bottomrule
\end{tabular}
\end{table*}

\begin{table*}[!t]\small
  \caption{Comparison with SOTA baseline dataset distillation methods on the Tiny-ImageNet dataset. In the first column, we present the accuracy (\%) achieved by various architectures on the full Tiny-ImageNet dataset. Our method significantly outperforms all compared baseline methods, as demonstrated in the table, even without the use of Dynamic Fine-Tuning (DFT). Incorporating DFT leads to a marked improvement in our method's accuracy. The table highlights the \textbf{highest accuracy in bold} and \underline{underlines the second-highest accuracy}. For the SCDD~\citep{SCDD} and GVBSM~\citep{shao2023generalized} methods, we list the results reported in the original papers.
  }
  \label{tab:tiny_imagenet_res}
  \centering
  \begin{tabular}{cc|cccc}
    \toprule

       \multirow{2}{*}{  Architecture}&  \multirow{2}{*}{  Method}   &  \multicolumn{4}{c}{  IPC }   \\
       \cmidrule(r){3-6}
       & & 1 & 10 & 50 & 100 \\
    \midrule
    \multirow{6}{*}{  ResNet-18 (59.6)} & SRe$^{2}$L & 2.62$\pm$0.1 & 16.1$\pm$0.2 & 41.1$\pm$0.4 &49.7$\pm$0.3  \\
    
    & SCDD &  - & 31.6$\pm$0.1 & 45.9$\pm$0.2 & -\\
    & GVBSM &  - & 47.6$\pm$0.3 & 51.0$\pm$0.4 & -\\
    & RDED &  9.7$\pm$0.4 & 41.9$\pm$0.2 & \textbf{58.2$\pm$0.1} & 59.1$\pm$0.1 \\

    & \name (Ours)  & \underline{16.5$\pm$0.2} & \underline{49.4$\pm$0.1} & 56.7$\pm$0.1& \underline{59.2$\pm$0.1}\\
    & \name + DFT (Ours)  & \textbf{21.2$\pm$0.1} & \textbf{51.1$\pm$0.1}& \underline{56.9$\pm$0.1} & \textbf{59.4$\pm$0.1}\\

     \midrule
    \multirow{5}{*}{  ResNet-50 (62.8)} & SRe$^{2}$L & 2.0$\pm$0.4 & 15.5$\pm$0.5&42.2$\pm$0.5&51.2$\pm$0.4    \\
    
    & GVBSM &  - & 48.7$\pm$0.2 & 52.1$\pm$0.3 & -\\
    & RDED &  8.1$\pm$0.3& 45.3$\pm$0.2& \textbf{61.6$\pm$0.3}& \textbf{62.6$\pm$0.1} \\
    & \name (Ours) & \underline{14.6$\pm$0.3} & \underline{53.4$\pm$0.1} & 59.8$\pm$0.2 & 62.0$\pm$0.2\\
    & \name + DFT (Ours) & \textbf{19.9$\pm$0.2} & \textbf{54.1$\pm$0.1} & \underline{ 60.9$\pm$0.2}& \underline{62.2$\pm$0.2} \\
    
     \midrule
    \multirow{4}{*}{  ResNet-101 (67.0)} & SRe$^{2}$L & 1.9$\pm$0.1 & 14.6$\pm$1.1& 42.5$\pm$0.2 & 51.5$\pm$0.3    \\
    & GVBSM &  - & 48.8$\pm$0.4 & 52.3$\pm$0.1 & -\\
    & RDED  &3.8$\pm$0.1& 22.9$\pm$3.3 & 41.2$\pm$0.4 & 65.2$\pm$1.1 \\
    & \name (Ours) &  \underline{13.2$\pm$0.2} & \underline{55.5$\pm$0.3} & \underline{63.2$\pm$0.2} & \underline{66.4$\pm$0.2} \\
    & \name + DFT (Ours) &  \textbf{19.4$\pm$0.2} & \textbf{56.3$\pm$0.2} & \textbf{64.1$\pm$0.2} & \textbf{67.0$\pm$0.1} \\

    \midrule
    \multirow{4}{*}{  MobileNet-V2 (45.2)} & SRe$^{2}$L & 2.0$\pm$0.3 & 7.3$\pm$0.2&19.5$\pm$0.4& 22.7$\pm$0.6    \\
    & RDED & 4.1$\pm$0.3 &27.4$\pm$0.3&40.1$\pm$0.2&42.6$\pm$0.3  \\
    & \name (Ours) & \underline{5.8$\pm$0.2} & \underline{34.8$\pm$0.2}  &\underline{ 42.2$\pm$0.1} & \underline{44.6$\pm$0.2} \\
    & \name + DFT (Ours)  & \textbf{5.9$\pm$0.3} & \textbf{36.6$\pm$0.2} & \textbf{43.6$\pm$0.1} & \textbf{45.0$\pm$0.3} \\

    \midrule
    \multirow{4}{*}{  EfficientNet-B0 (41.6)} & SRe$^{2}$L & 1.0$\pm$0.3 & 7.8$\pm$0.4 &17.5$\pm$0.7&20.9$\pm$0.3    \\
    & RDED &1.3$\pm$0.1&18.3$\pm$0.4& 38.2$\pm$0.3&40.4$\pm$0.2  \\
    & \name (Ours)  & \underline{7.5$\pm$0.1} & \underline{32.9$\pm$0.2} & \underline{40.4$\pm$0.2} & \underline{41.4$\pm$0.1}\\
    & \name + DFT (Ours)  & \textbf{9.0$\pm$0.1} & \textbf{33.5$\pm$0.2} & \textbf{41.2$\pm$0.3} & \textbf{42.7$\pm$0.1} \\
   
    \bottomrule
  \end{tabular}
\end{table*}

\textbf{Compiled Time and Memory Consumption.} 
Table~\ref{tab:synthesis_time_and_memory_consumption} presents the compiled time and memory consumption when utilizing a single RTX-4090 GPU on the ImageNet-1K dataset. Unlike SRe$^{2}$L, which consumes substantial resources, \name significantly reduces both compiled time and memory usage. Specifically, \name cuts the compiled time down to 8.67 seconds for Deit-S and 10.72 seconds for LV-ViT-S, while maintaining peak memory usage below 7 GB for Deit-S and slightly above 8 GB for LV-ViT-S \citep{NEURIPS2021_9a49a25d}. Compared with RDED, \name demonstrates a competitive advantage by achieving a more balanced utilization of time and GPU memory, thereby presenting a resource-efficient solution for dataset distillation.

The high efficiency of \name is attained through a strategy of down-sampling images before their input into the ViT model. This approach not only reduces the computational load but also enables a more flexible allocation of GPU resources through adaptive resizing of mini-batches. This efficiency is primarily due to the memory demands in our distillation process, which occur mainly during the parallel extraction of key informative patches within a mini-batch. Furthermore, the optimization-free nature of \name means that the distillation time per image depends on the size of the pre-trained ViT model used.


\begin{table*}[!t]\small 
  \caption{Compiled time and memory consumption on ImageNet-1K using a single RTX-4090 GPU. Time Cost is measured in seconds for generating 100 images simultaneously. Peak GPU memory usage is recorded for a batch size of 100, following the official SRe$^{2}$L~\citep{yin2024squeeze}  implementation. RDED-All indicates selection for all images in each category, whereas RDED only a random sample of 300 images per category.}
  \label{tab:synthesis_time_and_memory_consumption}
  \centering
  \begin{tabular}{cc|cc}
    \toprule
      Distillation Architecture & Method   & Time Cost (s)     & Peak Memory (GB) \\
    \midrule
    \multirow{3}{*}{  ResNet-18} & SRe$^{2}$L  & 211.32 & 9.14     \\
    & RDED & 3.99 & 1.57 \\
    & RDED-All & 26.34 & 8.63 \\

    \midrule
    \multirow{3}{*}{  MobileNet-V2} & SRe$^{2}$L  & 378.32 & 12.93     \\
    & RDED & 6.50 & 2.35 \\
    & RDED-All &31.27 & 11.06 \\

    \midrule
    \multirow{3}{*}{  EfficientNet-B0} & SRe$^{2}$L  & 441.24 & 11.92     \\
    & RDED & 7.32 & 2.34 \\
    & RDED-All & 37.83 & 10.96 \\
   \midrule
    \multirow{1}{*}{ Deit-S } & \name (Ours)  & 8.67 & 6.84     \\
   \multirow{1}{*}{ LV-ViT-S } & \name (Ours)  & 10.72 & 8.57     \\
    \bottomrule
  \end{tabular}
\end{table*}

\begin{table}
\centering
\caption{Comparative analysis of the accuracy and computational cost (measured in FLOPs) of training Deit-S on original versus downsampled images of ImageNet-1K.}
\begin{tabular}{c|cc}
\hline
    Resolutions & 224$\times$224 & 112$\times$112 \\
    \hline
    Accuracy & 79.8\% & 73.3\%  \\
    FLOPs & 4.60G & 1.10G \\
\hline
\end{tabular}

\label{tab:com_cost}
\end{table}

\textbf{Scaling up to Higher Resolutions.}
When the input resolution of ViT is expanded from \(224 \times 224\) to \(288 \times 288\), under the same hyperparameters, we evaluate the accuracy of compiled images using ResNet-18 and MobileNet-v2 on the ImageNet-1K dataset, as shown in Table~\ref{tab:higher_resolution}. We discover that despite increasing the resolution of the image input to ViT from \(224 \times 224\) to \(288 \times 288\), there is a slight decrease in accuracy. 
This phenomenon could be attributed to two factors. Firstly, a larger image resolution makes it more difficult to locate targets within the image, potentially leading to a decrease in the accuracy of the compiled dataset. Secondly, when training the validation model from scratch, all images are resized to the resolution of \(224 \times 224\). Reducing a higher-resolution image to this lower standard may result in more significant information loss.

\textbf{Impact of $\eta$ on performance.} 
Table~\ref{tab:blancing_factor} presents the accuracy of FocusDD across varying $\eta$ values. A smaller $\eta$ (e.g., $\eta$=0) denotes that representative images are selected based exclusively on the ViT model's prediction confidence scores, with subsequent target area selection guided by the attention scores of these images. Conversely, a larger $\eta$ (e.g., $\eta$=100) implies that representative images are chosen solely based on the highest attention area scores, followed by target area localization using the same attention scores. We adopt a moderate $\eta$ value of 30, which balances the representativeness of the images with the importance of their target areas, thereby achieving optimal accuracy.

\textbf{The advantages of downsampling.} The FocusDD synthetic dataset uses downsampled images to locate target regions for the following reasons: (1) Significant computational savings: As shown in Table~\ref{tab:com_cost}, downsampling reduces FLOPs by 4.2 times. (2) Facilitates dataset synthesis: It allows us to directly select low-resolution background images from the downsampled images to synthesize the final distilled image.

\begin{figure*}[h]
    \centering
    \begin{minipage}[t]{0.32\textwidth}
        \centering
        \includegraphics[width=\textwidth]{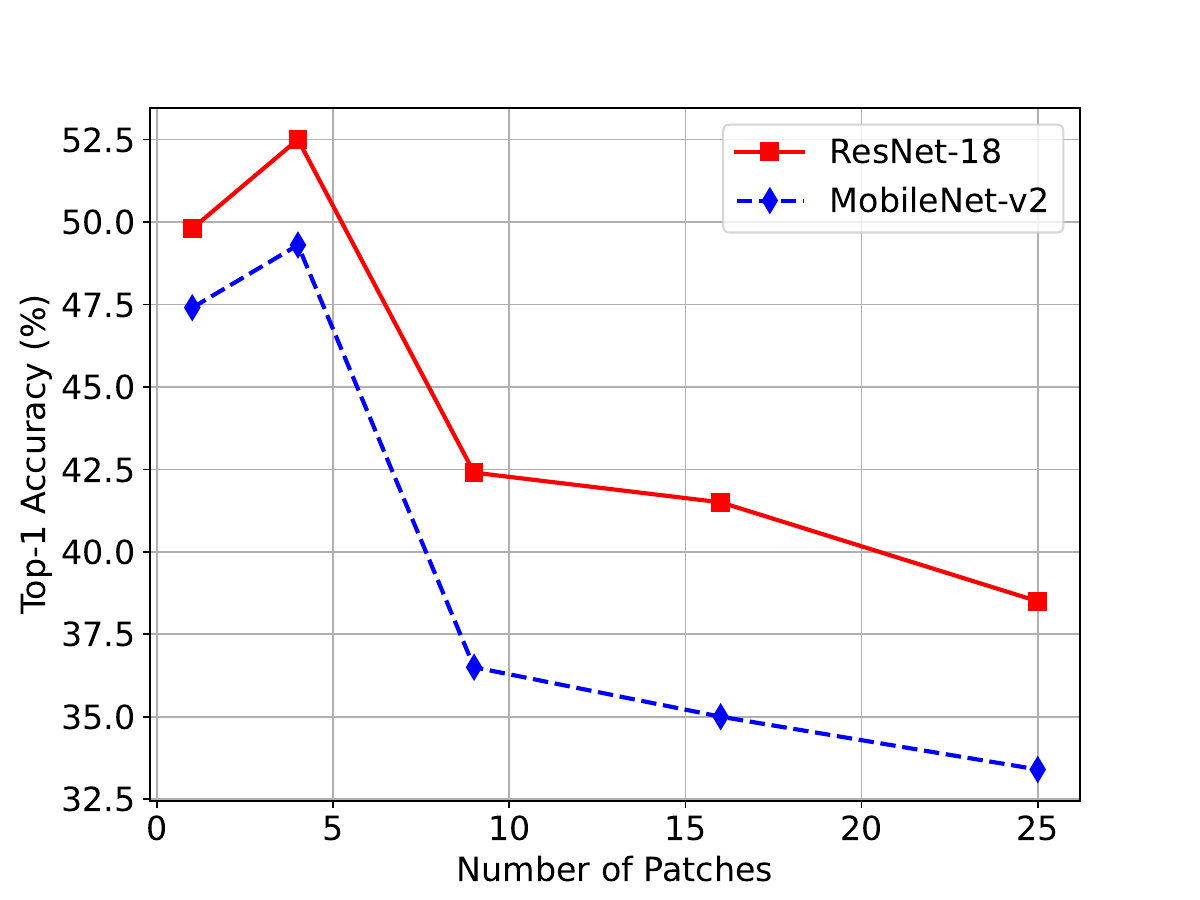} 
        \caption{The impact of the number of patches in each distilled image on accuracy. }
        \label{fig:patch_num}
    \end{minipage}
    \hfill
    \begin{minipage}[t]{0.32\textwidth}
        \centering
        \includegraphics[width=\textwidth]{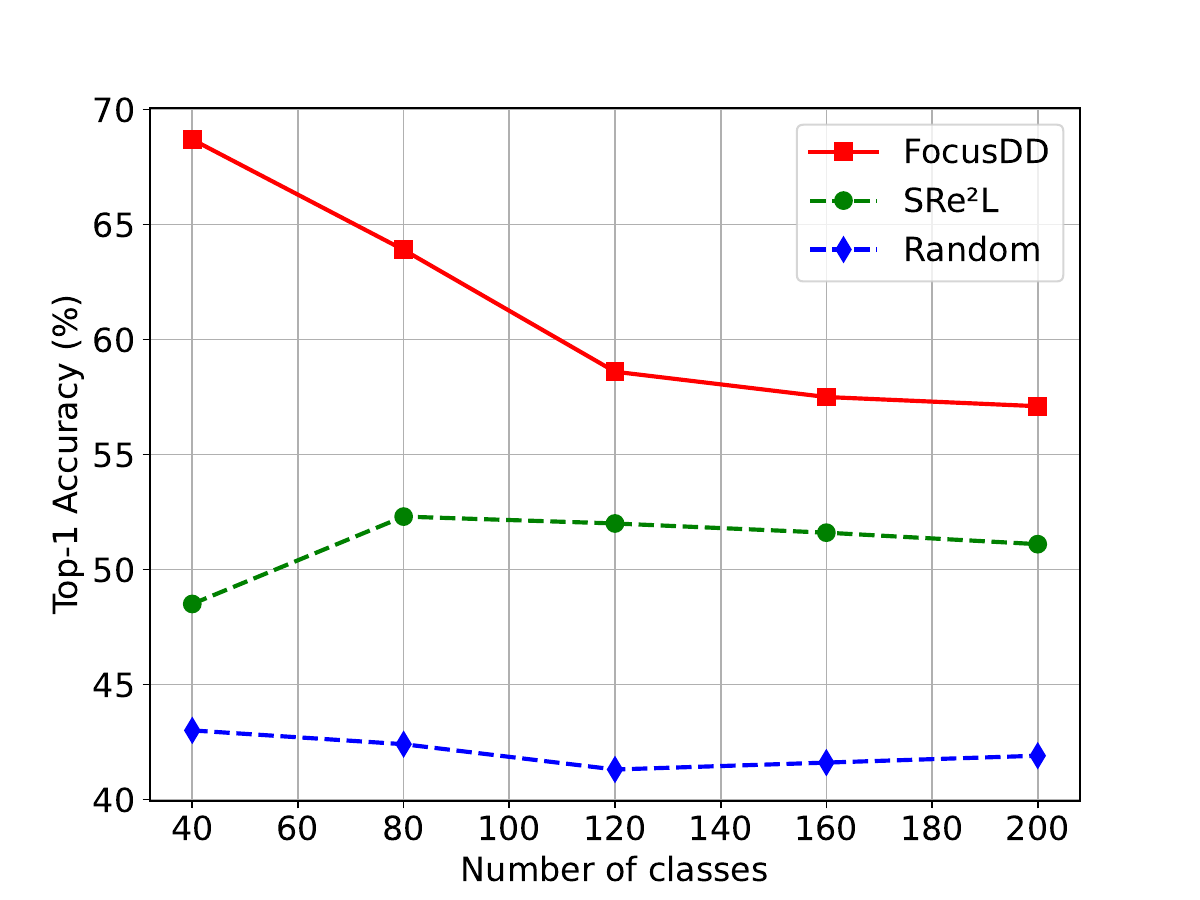} 
        \caption{5-step class-incremental learning on Tiny-ImageNet.}
        \label{fig:cl}
    \end{minipage}
    \hfill
    \begin{minipage}[t]{0.32\textwidth}
        \centering
        \includegraphics[width=\textwidth]{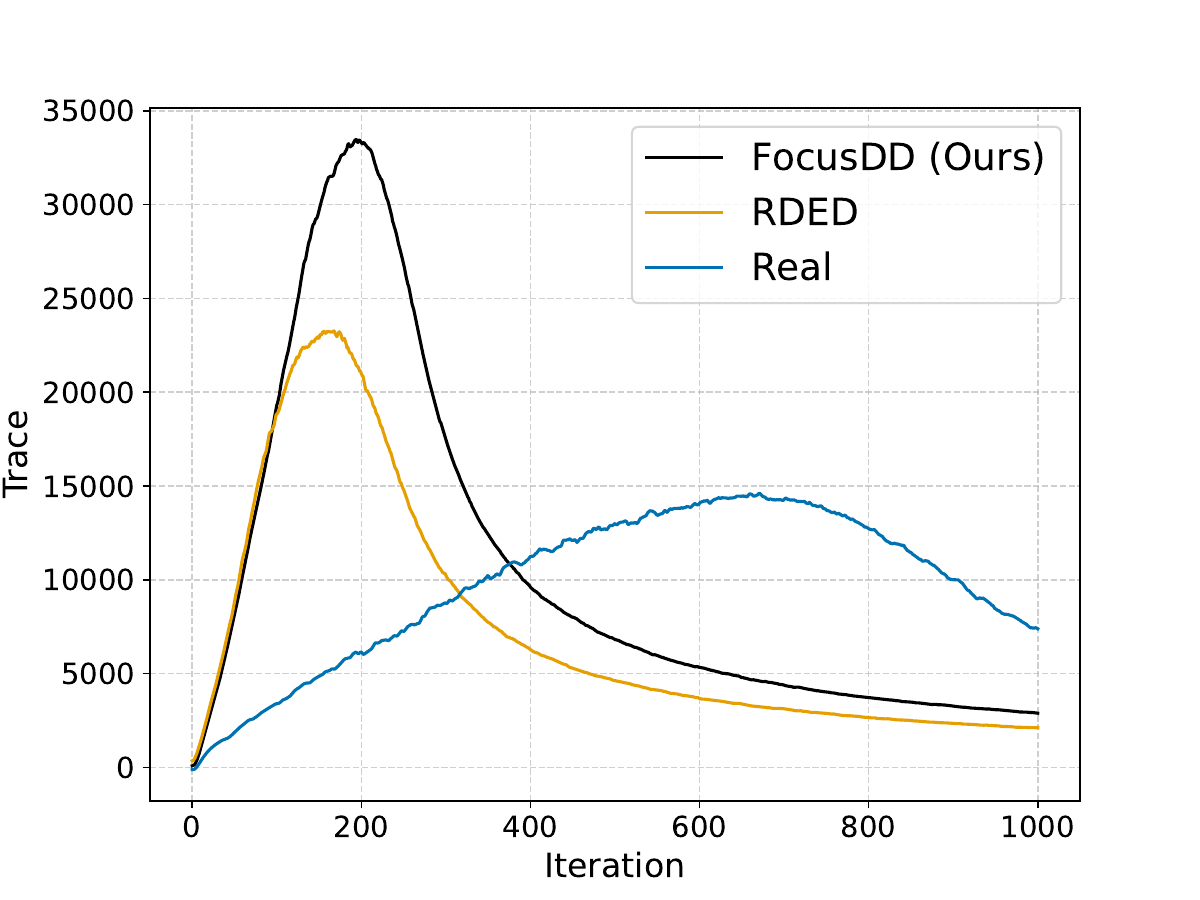} 
        \caption{Curvature of loss landscapes with real-vs-distilled data.}
        \label{fig:hessian}
    \end{minipage}

\end{figure*}

\textbf{Impact of the number of patches on performance.} Fig.~\ref{fig:patch_num} illustrates the impact of the number of patches in synthetic images on performance. We observed that as the number of patches increases, performance gradually decreases. This is because more patches reduce the resolution of each patch, making it difficult to accurately locate the target. Conversely, when the number of patches is 1, although the resolution is higher, the lack of diversity in the synthetic dataset leads to reduced performance. Considering these factors, we set the default number of patches to 4 to achieve optimal accuracy.

\textbf{Applications of synthetic datasets in continuous learning.} In Fig.~\ref{fig:cl}, we used ResNet18 and performed a 5-step validation on TinyImageNet to demonstrate FocusDD's performance in continual learning. The results show that FocusDD consistently surpasses the random baseline and matches or slightly exceeds SRe$^{2}$L~\citep{yin2024squeeze} as the number of classes increases from 40 to 200. This highlights its effectiveness in maintaining high accuracy and robustly adapting to new classes.

\textbf{Comparison of learning efficiency.} Fig.~\ref{fig:hessian} clearly shows the practical results of our attention-based approach, with FocusDD demonstrating higher learning efficiency compared to RDED. The higher Hessian matrix~\citep{yang2024dataset} trace values indicate that FocusDD not only adapts faster to new data but also absorbs basic data features more deeply, which is crucial for achieving high generalization in complex tasks.

\begin{table}\small 
\centering
\caption{When the input resolution for ViT is increased from $224\times 224$ to $288\times 288$, we evaluate the accuracy of the compiled images generated by \name. All accuracies were obtained after training for 1000 epochs on their respective datasets.} 
\label{tab:higher_resolution}
\begin{tabular}{c|c|c|c|c}

\toprule
Architec- &    \multicolumn{4}{c}{  IPC } \\
\cmidrule(r){2-5}
ture & 1& 10 & 50 & 100 \\
\midrule
R18 & 10.7$\pm$0.2  &52.5$\pm$0.1 & 63.1$\pm$0.1 & 68.0$\pm$0.2\\
R18$\#$288  & 9.6$\pm$0.2 &52.6$\pm$0.1 & 64.0$\pm$0.1 & 67.7$\pm$0.2\\
\midrule
Mv2 & 8.4$\pm$0.1 & 49.5$\pm$0.1 & 61.6$\pm$0.3 & 66.0$\pm$0.1\\
M2$\#$288  & 7.7$\pm$0.1 & 50.1$\pm$0.1 & 61.5$\pm$0.2 & 64.2$\pm$0.2\\
\bottomrule
\end{tabular}
\end{table}

\section{Theoretical Analysis}
\label{theory}

\subsection{Background and Definitions}

To analyze how the dataset distillation with an attention-based region selection affects the generalization ability of models on a testing dataset, we employ Rademacher Complexity \cite{yin2019rademacher} as a theoretical framework. We first present the setup and the analysis of our proposed \name method, followed by the empirical validation and the insights.

\textbf{Original Dataset \( D \).}
The original dataset, denoted as \( D \), consists of \( \vert D \vert \) samples, represented by \( \{\bm{x}_i\}_{i=1}^{\vert D \vert} \).

\textbf{Distilled Dataset \( D' \).}
The distilled dataset, \( D' \), is created by merging \( m \) samples from \( D \) based on key regions identified by an attention mechanism such as a Vision Transformer (ViT) and \( n \) samples with background information. This results in \( D' \) samples, \( \{\tilde{\bm{x}}_i\}_{i=1}^{ \vert D' \vert} \), where \( \vert D' \vert < \vert D \vert \).



\textbf{Rademacher Complexity.}
Rademacher Complexity measures the capacity of a class of functions to fit random noise, providing a metric for the complexity and generalization capability of hypothesis classes:
\[
\hat{R}_D(\mathcal{H}) = \mathbb{E}_\sigma \left[ \sup_{h \in \mathcal{H}} \frac{1}{\vert D \vert} \sum_{i=1}^{\vert D \vert} \sigma_i h(\bm{x}_i) \right],
\]
where \( \sigma_i \) are independent random variables taking values \( +1 \) or \( -1 \) with equal probability.
We apply this metric when evaluating the distilled datasets because it can provide insight into whether the distillation process preserves the richness of the hypothesis space or if it overly simplifies the dataset, potentially losing important variances needed for higher generalization.

\subsection{Impact of Dataset Distillation of \name}
\label{appendix:theory_2}

For the distilled dataset \( S' \), the Rademacher Complexity becomes:
\[
\hat{R}_{D'}(\mathcal{H}) = \mathbb{E}_\sigma \left[ \sup_{h \in \mathcal{H}} \frac{1}{\vert D' \vert} \sum_{i=1}^{\vert D' \vert} \sigma_i h(\tilde{\bm{x}}_i) \right].
\]

Each distilled data instance \( \tilde{\bm{x}}_i = \text{concatenate}(\{\bm{x}_{q}^\star\}_{q=1}^m, \{\bm{x}_{l}^\prime\}_{l=1}^n) \), where \( \bm{x}^\star \) represents the key sub-region data and \(\bm{x}^\prime \) means the down-scaled low resolution data with backgound information.

Note that the term \( 1 / \vert D^{'} \vert \) determines the scaling of the sum of fits to random labels (noise) in the Rademacher Complexity formula. When analyzing a dataset that has undergone distillation to produce \( D' \), where each sample \( \tilde{\bm{x}}_i \)  aggregates the informational content of multiple samples from the original dataset, the actual number of samples \( \vert D' \vert \) might not accurately reflect the dataset's complexity. 
Instead, the Efficient Sample Size (ESS)  \citep{elvira2022rethinking} is applied to represent the number of independent observations in a dataset that would provide the same amount of information as the actual dataset, which can be noted as \( \vert D'_{\text{eff}} \vert \). If \( \vert D'_{\text{eff}} \vert \) represents a more accurate measure of the independent information content in \( D' \), the complexity measure can be adjusted to:
   \[
   \hat{R}_{D'}(\mathcal{H}) = \mathbb{E}_\sigma \left[ \sup_{h \in \mathcal{H}} \frac{1}{\vert D'_{\text{eff}} \vert} \sum_{i=1}^{\vert D' \vert} \sigma_i h(\tilde{\bm{x}}_i) \right].
   \]
This adjustment recognizes that the effective diversity and informational independence in \( D' \) might be greater than simply counting \( \vert D' \vert \), hence potentially leading to a more accurate estimation of how the hypothesis class \( \mathcal{H} \) will perform. 

The complexity induced by each new sample \( \tilde{\bm{x}}_i \) can reduce the variance among samples, as they inherently represent a more uniform distribution of the key features and contexts of the original dataset. The formula for Rademacher Complexity has to consider the effective sample size \( \vert D'_\text{eff} \vert \) that accounts for this aggregation:
\[
\vert D'_\text{eff} \vert = \vert D' \vert \times (m*\gamma + n*\beta),
\]
\noindent where $\gamma$ and $\beta$ represent the degression parameters due to selecting only the key regions or using down-scaled data, which range from 0 to 1. The setting $\gamma = \beta = 1$ means that we naively concatenate \(m+n\) original data instances.

\begin{table*}[t]
\centering
\caption{\textbf{Rademacher Complexity Comparison with the same IPC.} Na\"ive denotes randomly selecting \( \vert D' \vert \) samples from the original dataset, RDED concatenates $(m+n)$ random sub-region samples. \(\tau\) is a regression parameter due to selecting only the sub-regions.}
\label{tab:Rademacher}
\begin{tabular}{c|c|c}
\toprule
Method & \(\tilde{\bm{x}}_i\) & \( \vert D'_\text{eff} \vert\)  \\
\midrule
Na\"ive & \(\bm{x}_i\) & \( \vert D' \vert\)  \\
RDED &  \( \text{concatenate}(\{\bm{x}_{j}^{\text{rand}}\}_{j=1}^{m+n}) \) & \( \vert D' \vert \times (m+n)*\tau\) \\
\name (Ours) & \( \text{concatenate}(\{\bm{x}_{q}^\star\}_{q=1}^m, \{\bm{x}_{l}^\prime\}_{l=1}^n) \) & \( \vert D' \vert \times (m*\gamma + n*\beta)\) \\
\bottomrule
\end{tabular}
\end{table*}

\begin{figure*}[h]
    \centering
    \begin{subfigure}[b]{0.3\linewidth}
        \includegraphics[width=\linewidth]{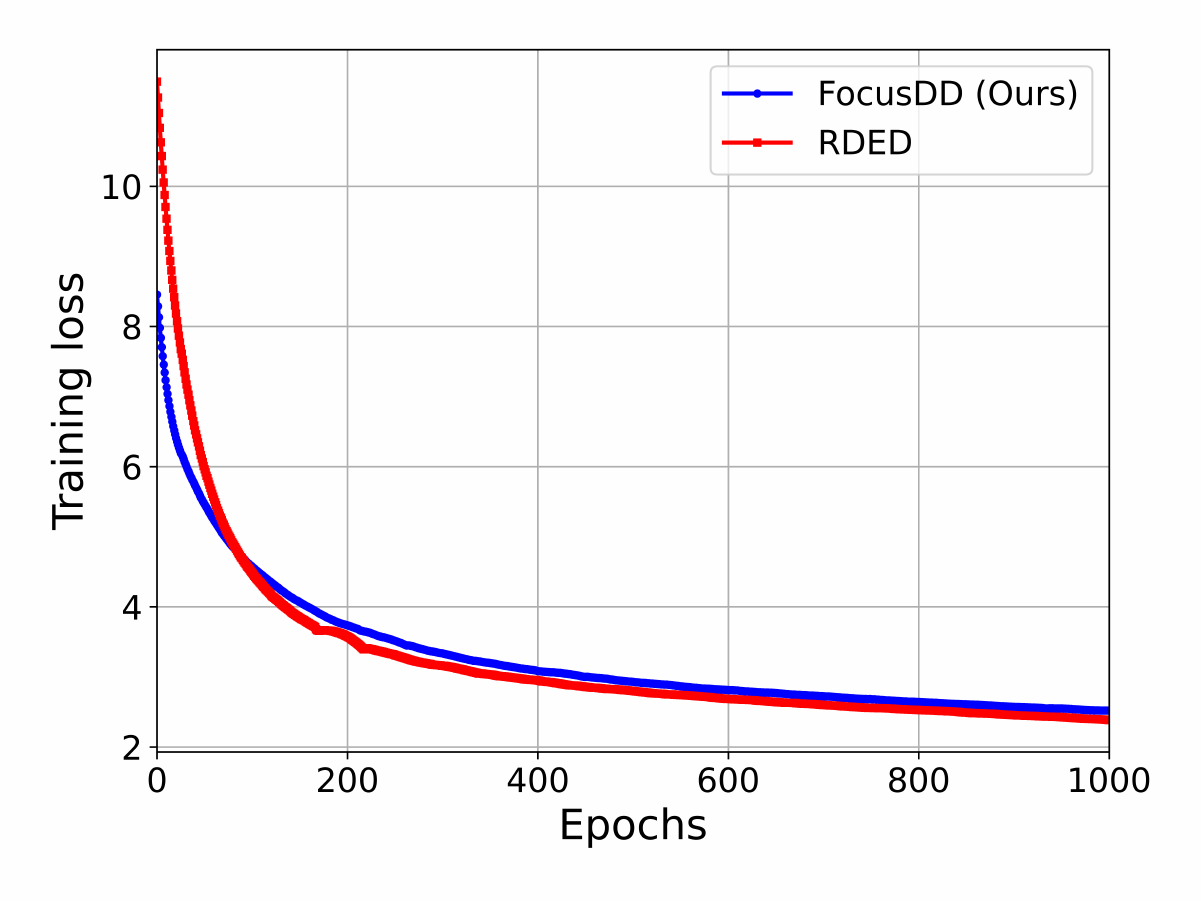}
        \caption{}
        \label{fig:subfig1}
    \end{subfigure}
    \hfill
    \begin{subfigure}[b]{0.3\linewidth}
        \includegraphics[width=\linewidth]{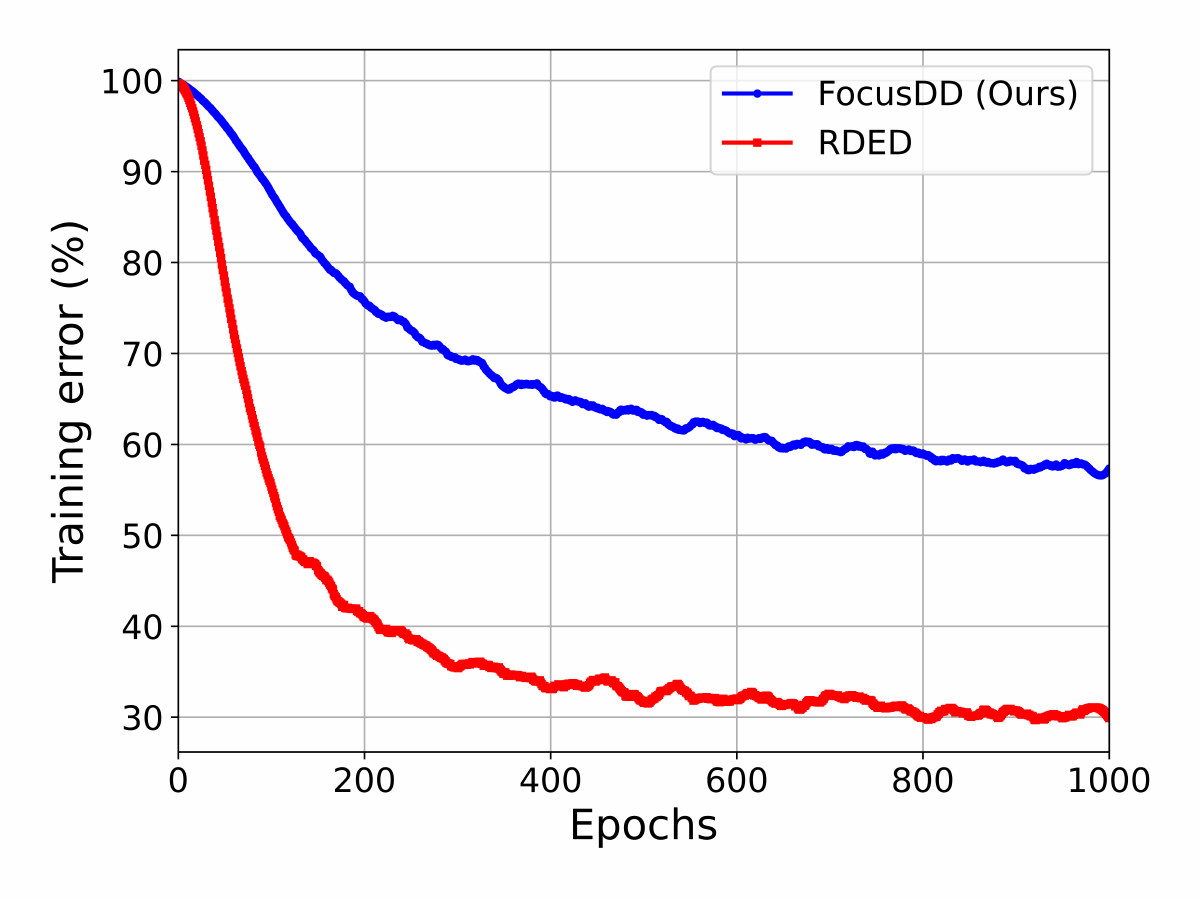}
        \caption{}
        \label{fig:subfig2}
    \end{subfigure}
    \hfill
    \begin{subfigure}[b]{0.3\linewidth}
        \includegraphics[width=\linewidth]{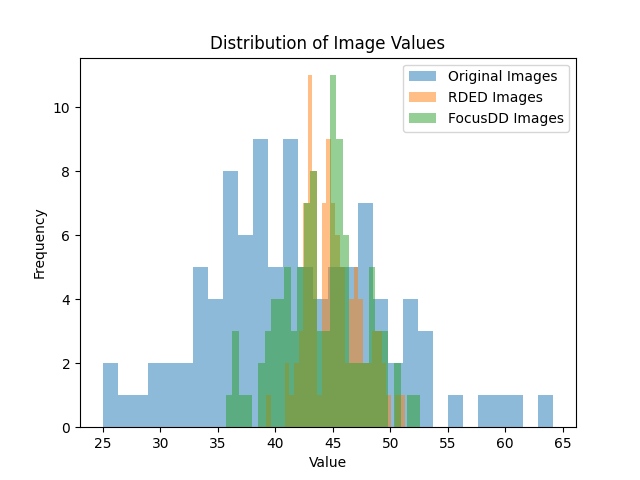}
        \caption{}
        \label{fig:subfig3}
    \end{subfigure}
    \caption{The diversity of the compiled dataset is assessed by analyzing the training loss and accuracy metrics on the compiled image training set. (a) Training loss. (b) Training error. (c) The signal-to-noise distribution of images within the same category for the full dataset and those distilled by RDED and \name. Training loss on compiled images. All methods employ MobileNet-v2 and are executed on the ImageNet-1K dataset with IPC = 10. }
    \label{fig:iversity_res}
\end{figure*}

Similarly, we can determine \(\tilde{\bm{x}}_i\) and \(\vert D'_\text{eff} \vert\) for two baseline methods as shown in Table.~\ref{tab:Rademacher}: Na\"ive and RDED \citep{sun2024diversity}. A higher \(\vert D'_\text{eff} \vert\) indicates that each sample in \( D' \) contains more "independent-like" information than initially apparent, suggesting that \( D' \) may exhibit a lower Rademacher Complexity than expected if assessed solely based on \(\vert D' \vert\). Generally, a lower Rademacher Complexity correlates with better generalization capabilities, indicating that models trained on \( D' \) might generalize better than anticipated based solely on \(\vert D' \vert\).
This enhanced generalization is why RDED and our proposed method significantly outperform the Na\"ive approach, which relies on random sample selection.

Our method employs strategies to achieve a larger \(\vert D'_\text{eff} \vert\) than RDED. Our realism score \(s_i^{\text{real}}\) combines the predictive confidence score and the maximum attention region score. When selecting samples, it does not only consider the information richness of the samples but also the information density of the target regions within these samples. Together, these factors improve \(\vert D'_\text{eff} \vert\) and enhance generalization capabilities, as confirmed by the results in Table~\ref{tab:patch_number_effect} for \(m \neq 0\), which reflect the combined effect of both strategies.


Our method, \name, is also designed to reduce model complexity within the Hypothesis Space. Richer samples may enable the functions \( h \) in \( \mathcal{H} \) to be less complex, as each sample encompasses a broader range of information, potentially simplifying the learning problem. This hypothesis is supported by the results in Table~\ref{tab:imagenet_res}, which demonstrate that simpler backbone models using \name data achieve outcomes comparable to those of more complex models.



\textbf{Quantifying the Diversity and SNR of Synthetic Images.}
We employ the method outlined in \citet{lopes2021tradeoffs} to assess the diversity of compiled images. According to \citet{lopes2021tradeoffs}, greater dataset diversity presents more challenges for the training process to converge, often resulting in larger loss values and longer training times. Fig.~\ref{fig:iversity_res}(a) compares the training loss of our method with the SOTA method RDED \citep{sun2024diversity} on compiled datasets. Initially, our \name method starts with lower loss values but ends with higher losses than RDED after training. Moreover, Fig.~\ref{fig:iversity_res}(b) illustrates significant differences in accuracy tests on the training dataset, indicating that images synthesized using our method are more diverse and thus harder to train. This observation aligns with the conclusions in \citet{lopes2021tradeoffs}, confirming that our approach generates more diverse compiled images, making the training process more challenging but potentially leading to more robust models.

Fig.~\ref{fig:iversity_res}(c) illustrates the distribution of signal-to-noise ratios (SNR)\footnote{We applied a 3$\times$3 Laplacian kernel to filter the images to extract their high-frequency components. Then, we calculated the sum of the absolute values of the convolution results between the image and this matrix, using this to estimate the standard deviation of the noise. Finally, based on the definition of signal-to-noise ratio, we computed the SNR distribution for the entire dataset.} for the original dataset and datasets processed by two different distillation methods, within the same category. The SNR distribution of the original images is relatively concentrated, with most values ranging between 30 and 58. The SNR of images processed by RDED \citep{sun2024diversity} shifts to the right, primarily distributed between 42 and 50. In contrast, images processed by \name exhibit a wider SNR distribution, spanning from 36 to 53. Although the average SNR of RDED images is the highest at 45.1, the average SNR for \name images is 44.0, closer to the original dataset's average SNR of 41.7. This indicates that the \name method effectively enhances image quality while preserving the characteristics of the original data, thereby demonstrating superior balanced performance in practical applications.


\subsection{Remarks}
The proposed distillation method, \name, is expected to enhance generalization by utilizing more informative and representative samples. The associated reduction in Rademacher Complexity indicates a diminished capacity for fitting random noise, which typically suggests improved performance on unseen data.

The practical implementation may encounter challenges, such as increased computational overhead from processing larger \( \tilde{\bm{x}}_i \) values. Additionally, there is a risk of information redundancy if the parameters \( m \) and \( n \) are not optimally selected.

\section{Sample Visualizations of Synthetic Data}
\label{Visualization_Synthetic_Data}
\begin{figure*}[ht]
  \centering
  \includegraphics[width=\textwidth]{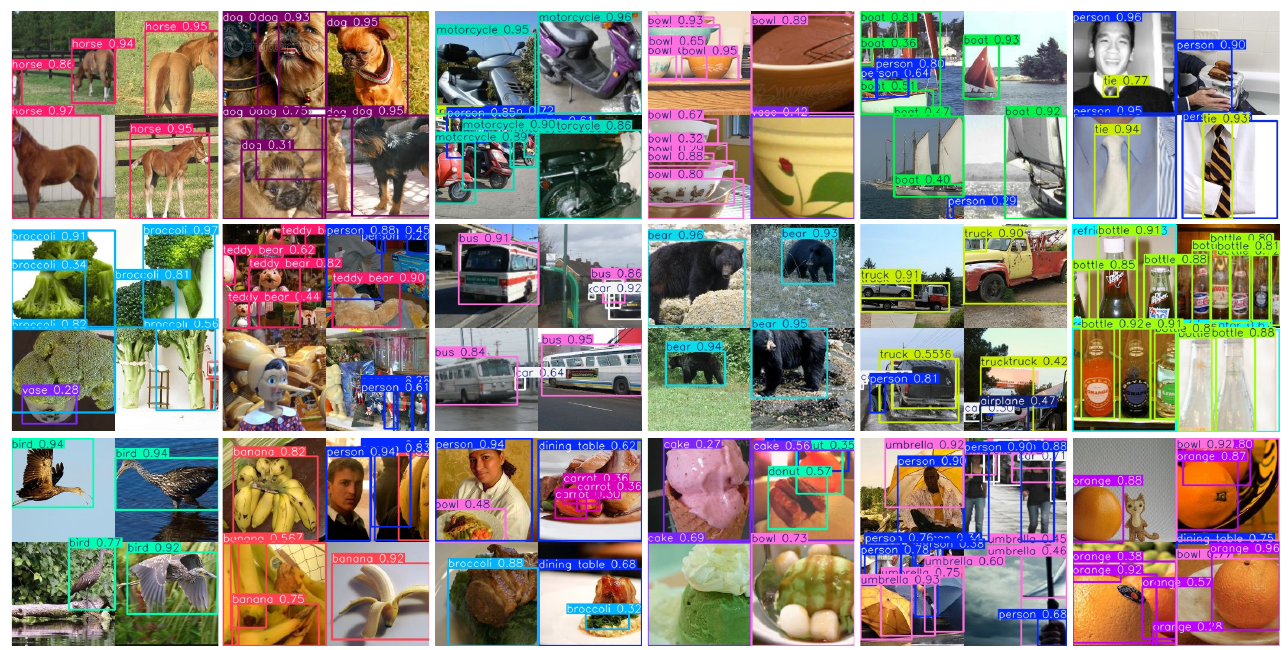}
\caption{Visualization examples of training samples for object detection generated by FocusDD.}

  \label{fig:object_detect}
\end{figure*}

\begin{figure*}[h]
  \centering
  \includegraphics[width=0.8\textwidth,height=1.2\linewidth]{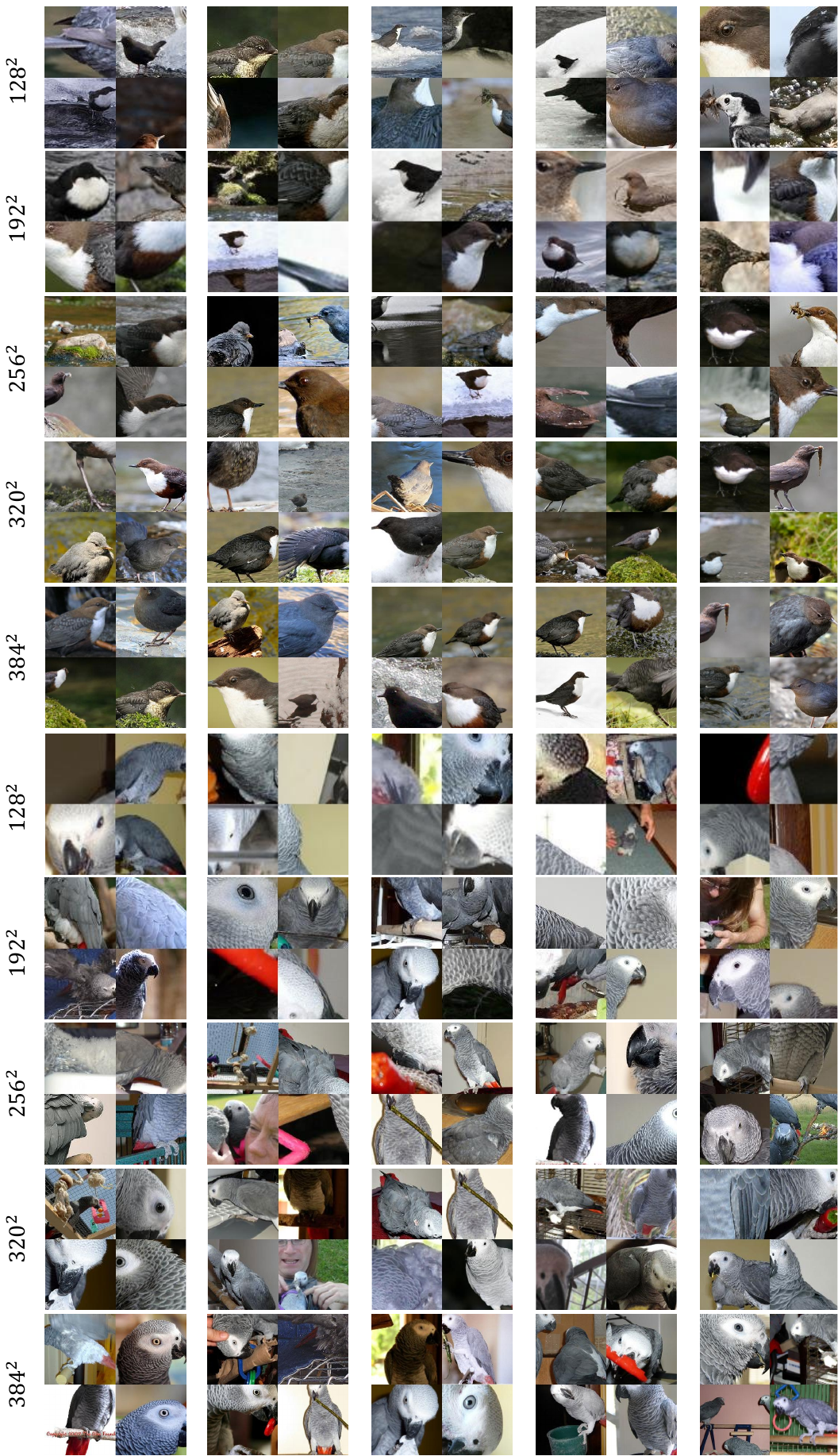}
\caption{\name compiled images with different resolutions on the ImageNet-1K dataset.  We can clearly see that as the resolution increases, each patch in the compiled image gradually expands from containing only a part of the target to including the entire target, thereby enhancing the accuracy of the image (Fig.~\ref{fig:adaptive_resolution_accuracy}).}

  \label{fig:adaptive_resolution}
\end{figure*}

\begin{figure*}[h]
  \centering
   \setlength{\abovecaptionskip}{0pt}
   \setlength{\belowcaptionskip}{0pt}
  \includegraphics[width=0.80\textwidth,height=1.3\linewidth]{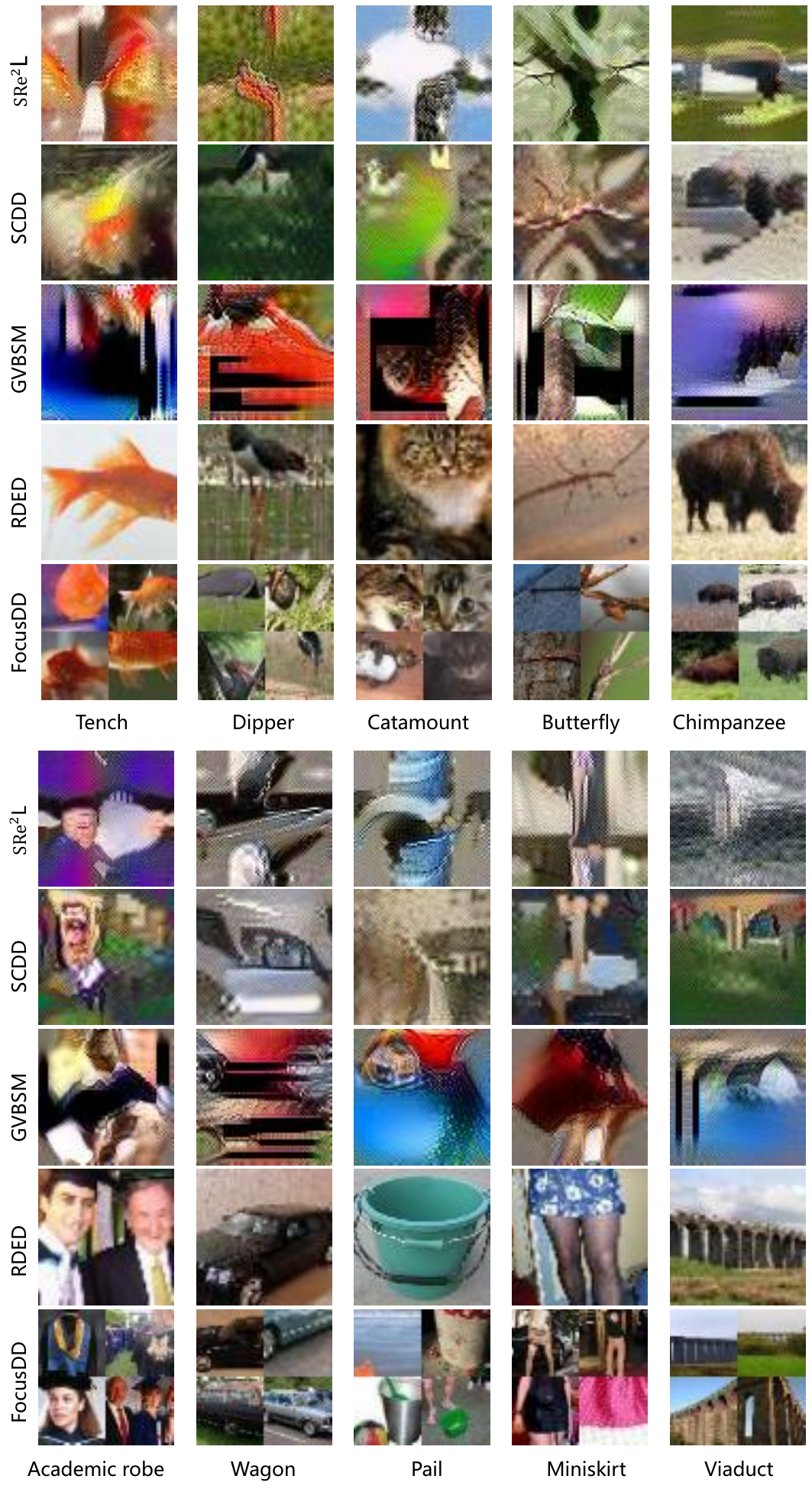}
\caption{Compiled data visualization on Tiny-ImageNet from SRe$^2$L \citep{yin2024squeeze}, SCDD \citep{SCDD}, GVBSM \citep{shao2023generalized}, RDED \citep{sun2024diversity} and \name.}

  \label{fig:tinyimagenet_visualization}
\end{figure*}

\begin{figure*}[h]
  \centering
   \setlength{\abovecaptionskip}{0pt}
   \setlength{\belowcaptionskip}{0pt}
  \includegraphics[width=0.8\textwidth,height=1.3\linewidth]{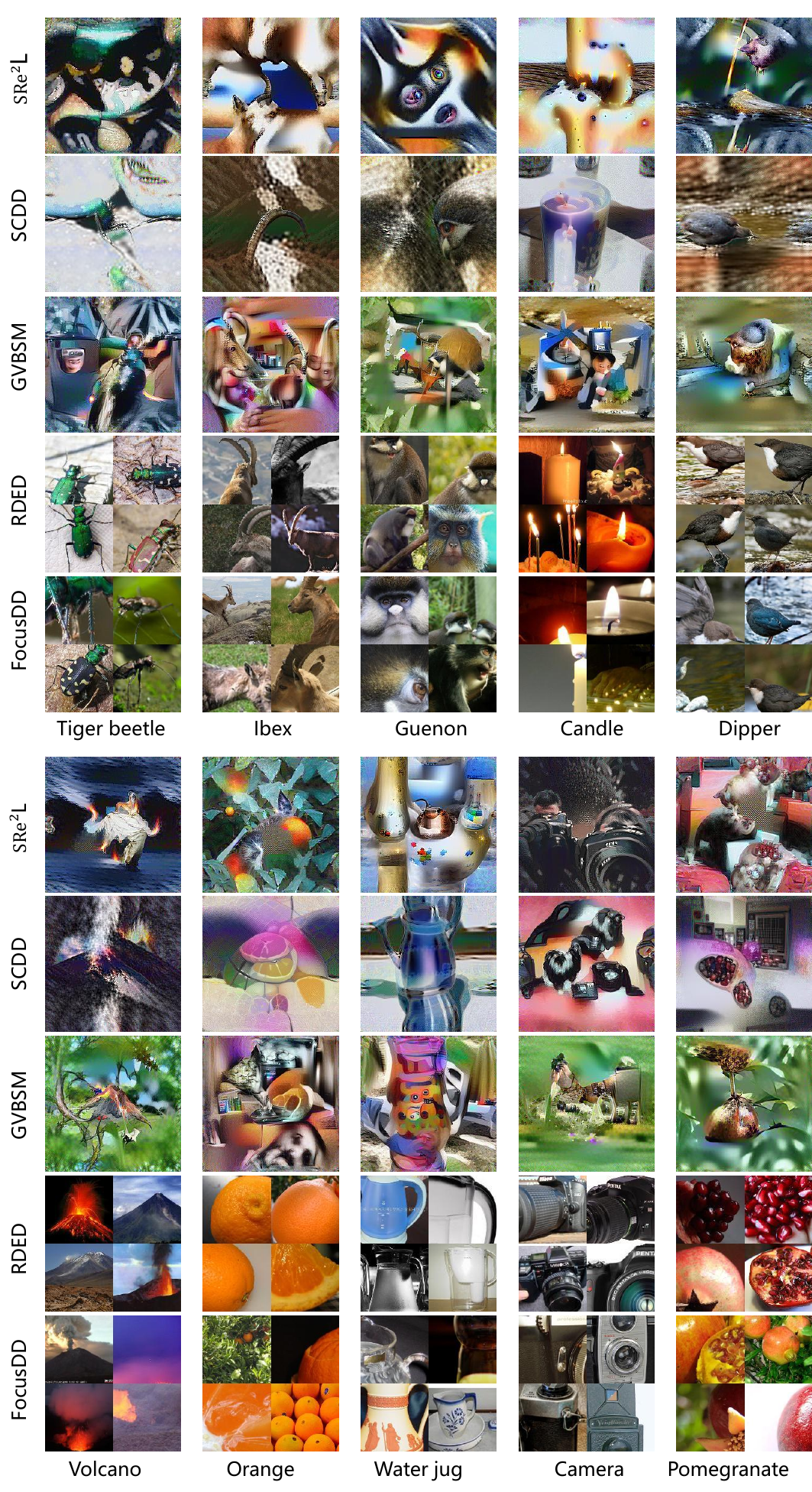}
\caption{Compiled data visualization on ImageNet-1K from SRe$^2$L \citep{yin2024squeeze}, SCDD \citep{SCDD}, GVBSM \citep{shao2023generalized}, RDED \citep{sun2024diversity} and \name.}

  \label{fig:imagenet_visualization}
\end{figure*}

\begin{figure*}[h]
  \centering
   \setlength{\abovecaptionskip}{0pt}
   \setlength{\belowcaptionskip}{0pt}
  \includegraphics[width=0.8\textwidth,height=1.3\linewidth]{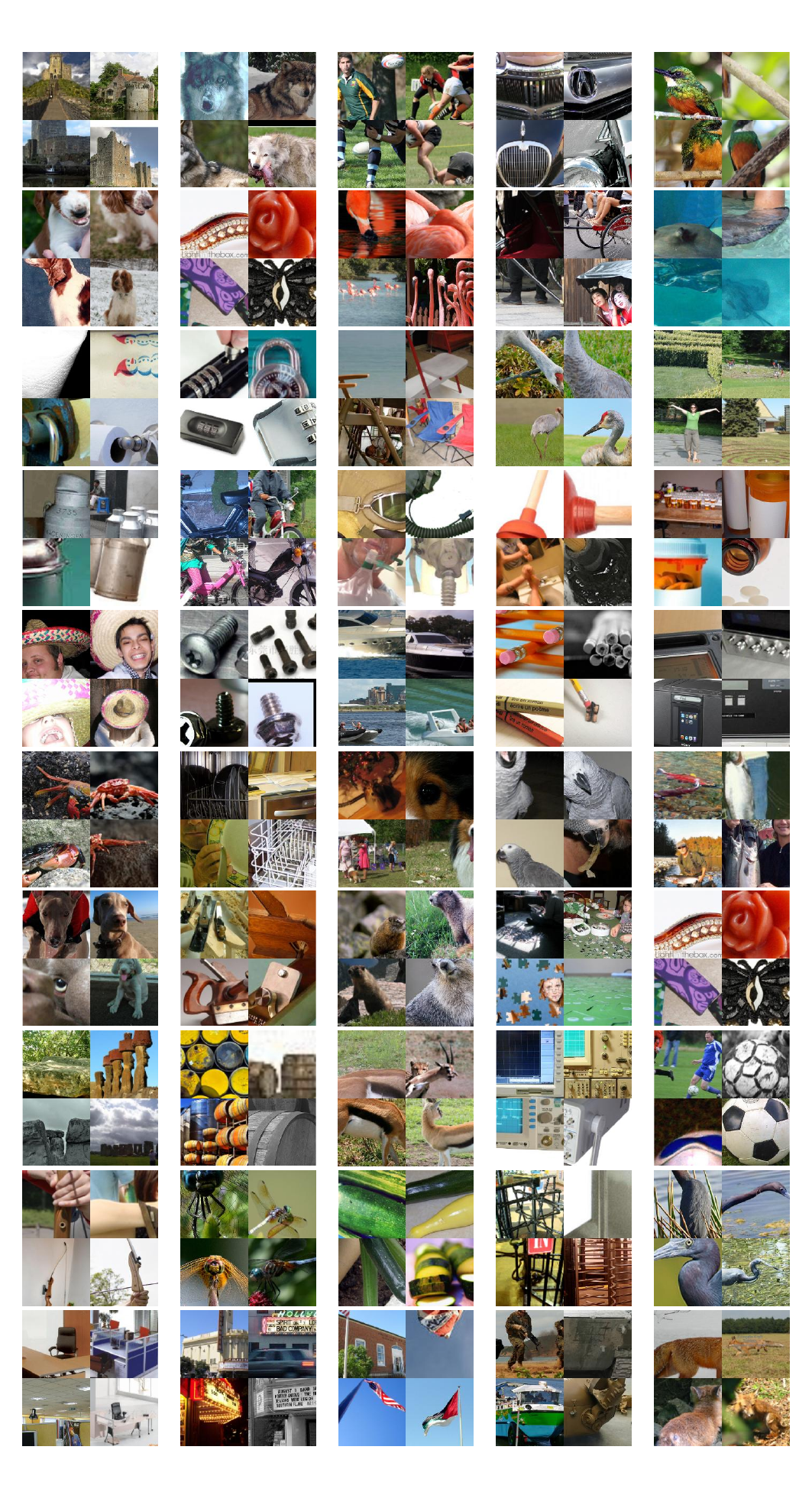}
\caption{Compiled data visualization on ImageNet-1K from \name.}

  \label{fig:more_visualization}
\end{figure*}

Fig.~\ref{fig:object_detect} presents visualization examples of object detection training samples generated by FocusDD. Fig.~\ref{fig:adaptive_resolution} further compares \name-compiled images at different resolutions, showing that as resolution increases, each image patch transitions from capturing only parts of objects to representing entire objects. This trend is quantified in Fig.~\ref{fig:adaptive_resolution_accuracy}, which also highlights a corresponding improvement in accuracy.  
In Fig.~\ref{fig:tinyimagenet_visualization}, we compare the Tiny-ImageNet samples compiled by SRe$^2$L \citep{yin2024squeeze}, SCDD \citep{SCDD}, GVBSM \citep{shao2023generalized}, RDED \citep{sun2024diversity}, and \name. To provide a more comprehensive perspective, Figs.~\ref{fig:imagenet_visualization} and \ref{fig:more_visualization} present visualizations of compiled samples from ImageNet-1K. Our compiled data, cropped directly from real image target areas, demonstrates superior realism in texture, shape, and detail compared to SRe$^2$L, SCDD, and GVBSM. Unlike RDED, our method incorporates a low-resolution background in the compiled images, enriching them with additional semantic information. These results collectively demonstrate the higher quality of our compiled data.

%% file: arxiv.bbl
\begin{thebibliography}{67}
\providecommand{\natexlab}[1]{#1}
\providecommand{\url}[1]{\texttt{#1}}
\expandafter\ifx\csname urlstyle\endcsname\relax
  \providecommand{\doi}[1]{doi: #1}\else
  \providecommand{\doi}{doi: \begingroup \urlstyle{rm}\Url}\fi

\bibitem[Bossard et~al.(2014)Bossard, Guillaumin, and Van~Gool]{bossard2014food}
L.~Bossard, M.~Guillaumin, and L.~Van~Gool.
\newblock Food-101--mining discriminative components with random forests.
\newblock In \emph{Computer vision--ECCV 2014: 13th European conference, zurich, Switzerland, September 6-12, 2014, proceedings, part VI 13}, pages 446--461. Springer, 2014.

\bibitem[Carion et~al.(2020)Carion, Massa, Synnaeve, Usunier, Kirillov, and Zagoruyko]{carion2020end}
N.~Carion, F.~Massa, G.~Synnaeve, N.~Usunier, A.~Kirillov, and S.~Zagoruyko.
\newblock End-to-end object detection with transformers.
\newblock In \emph{European conference on computer vision}, pages 213--229. Springer, 2020.

\bibitem[Cazenavette et~al.(2022)Cazenavette, Wang, Torralba, Efros, and Zhu]{cazenavette2022dataset}
G.~Cazenavette, T.~Wang, A.~Torralba, A.~A. Efros, and J.-Y. Zhu.
\newblock Dataset distillation by matching training trajectories.
\newblock In \emph{Proceedings of the IEEE/CVF Conference on Computer Vision and Pattern Recognition}, pages 4750--4759, 2022.

\bibitem[Cazenavette et~al.(2023)Cazenavette, Wang, Torralba, Efros, and Zhu]{cazenavette2023generalizing}
G.~Cazenavette, T.~Wang, A.~Torralba, A.~A. Efros, and J.-Y. Zhu.
\newblock Generalizing dataset distillation via deep generative prior.
\newblock In \emph{Proceedings of the IEEE/CVF Conference on Computer Vision and Pattern Recognition}, pages 3739--3748, 2023.

\bibitem[Chen et~al.(2023{\natexlab{a}})Chen, Lin, Li, Shen, Wu, Chao, and Ji]{chen2023cf}
M.~Chen, M.~Lin, K.~Li, Y.~Shen, Y.~Wu, F.~Chao, and R.~Ji.
\newblock Cf-vit: A general coarse-to-fine method for vision transformer.
\newblock In \emph{Proceedings of the AAAI Conference on Artificial Intelligence}, volume~37, pages 7042--7052, 2023{\natexlab{a}}.

\bibitem[Chen et~al.(2023{\natexlab{b}})Chen, Lin, Lin, Zhang, Chao, and Ji]{chen2023smmix}
M.~Chen, M.~Lin, Z.~Lin, Y.~Zhang, F.~Chao, and R.~Ji.
\newblock Smmix: Self-motivated image mixing for vision transformers.
\newblock In \emph{Proceedings of the IEEE/CVF International Conference on Computer Vision}, pages 17260--17270, 2023{\natexlab{b}}.

\bibitem[Chrabaszcz et~al.(2017)Chrabaszcz, Loshchilov, and Hutter]{chrabaszcz2017downsampled}
P.~Chrabaszcz, I.~Loshchilov, and F.~Hutter.
\newblock A downsampled variant of imagenet as an alternative to the cifar datasets.
\newblock \emph{arXiv preprint arXiv:1707.08819}, 2017.

\bibitem[Cui et~al.(2023)Cui, Wang, Si, and Hsieh]{cui2023scaling}
J.~Cui, R.~Wang, S.~Si, and C.-J. Hsieh.
\newblock Scaling up dataset distillation to imagenet-1k with constant memory.
\newblock In \emph{International Conference on Machine Learning}, pages 6565--6590. PMLR, 2023.

\bibitem[Deng et~al.(2009)Deng, Dong, Socher, Li, Li, and Fei-Fei]{deng2009imagenet}
J.~Deng, W.~Dong, R.~Socher, L.-J. Li, K.~Li, and L.~Fei-Fei.
\newblock Imagenet: A large-scale hierarchical image database.
\newblock In \emph{2009 IEEE conference on computer vision and pattern recognition}, pages 248--255. Ieee, 2009.

\bibitem[Deng and Russakovsky(2022)]{deng2022remember}
Z.~Deng and O.~Russakovsky.
\newblock Remember the past: Distilling datasets into addressable memories for neural networks.
\newblock \emph{Advances in Neural Information Processing Systems}, 35:\penalty0 34391--34404, 2022.

\bibitem[Dosovitskiy et~al.(2020)Dosovitskiy, Beyer, Kolesnikov, Weissenborn, Zhai, Unterthiner, Dehghani, Minderer, Heigold, Gelly, et~al.]{dosovitskiy2020image}
A.~Dosovitskiy, L.~Beyer, A.~Kolesnikov, D.~Weissenborn, X.~Zhai, T.~Unterthiner, M.~Dehghani, M.~Minderer, G.~Heigold, S.~Gelly, et~al.
\newblock An image is worth 16x16 words: Transformers for image recognition at scale.
\newblock In \emph{International Conference on Learning Representations}, 2020.

\bibitem[Du et~al.(2023)Du, Jiang, Tan, Zhou, and Li]{du2023minimizing}
J.~Du, Y.~Jiang, V.~Y. Tan, J.~T. Zhou, and H.~Li.
\newblock Minimizing the accumulated trajectory error to improve dataset distillation.
\newblock In \emph{Proceedings of the IEEE/CVF Conference on Computer Vision and Pattern Recognition}, pages 3749--3758, 2023.

\bibitem[Elvira et~al.(2022)Elvira, Martino, and Robert]{elvira2022rethinking}
V.~Elvira, L.~Martino, and C.~P. Robert.
\newblock Rethinking the effective sample size.
\newblock \emph{International Statistical Review}, 90\penalty0 (3):\penalty0 525--550, 2022.

\bibitem[Feldman and Zhang(2020)]{feldman2020neural}
V.~Feldman and C.~Zhang.
\newblock What neural networks memorize and why: Discovering the long tail via influence estimation.
\newblock \emph{Advances in Neural Information Processing Systems}, 33:\penalty0 2881--2891, 2020.

\bibitem[Forgy(1965)]{forgy1965cluster}
E.~W. Forgy.
\newblock Cluster analysis of multivariate data: efficiency versus interpretability of classifications.
\newblock \emph{biometrics}, 21:\penalty0 768--769, 1965.

\bibitem[Gontijo-Lopes et~al.(2021)Gontijo-Lopes, Smullin, Cubuk, and Dyer]{lopes2021tradeoffs}
R.~Gontijo-Lopes, S.~Smullin, E.~D. Cubuk, and E.~Dyer.
\newblock Tradeoffs in data augmentation: An empirical study.
\newblock In \emph{International Conference on Learning Representations}, 2021.

\bibitem[Gu et~al.(2024)Gu, Vahidian, Kungurtsev, Wang, Jiang, You, and Chen]{gu2023efficient}
J.~Gu, S.~Vahidian, V.~Kungurtsev, H.~Wang, W.~Jiang, Y.~You, and Y.~Chen.
\newblock Efficient dataset distillation via minimax diffusion.
\newblock In \emph{Proceedings of the IEEE/CVF Conference on Computer Vision and Pattern Recognition (CVPR)}, 2024.

\bibitem[Guo et~al.(2023)Guo, Wang, Cazenavette, Li, Zhang, and You]{guo2023towards}
Z.~Guo, K.~Wang, G.~Cazenavette, H.~Li, K.~Zhang, and Y.~You.
\newblock Towards lossless dataset distillation via difficulty-aligned trajectory matching.
\newblock \emph{arXiv preprint arXiv:2310.05773}, 2023.

\bibitem[He et~al.(2016)He, Zhang, Ren, and Sun]{he2016deep}
K.~He, X.~Zhang, S.~Ren, and J.~Sun.
\newblock Deep residual learning for image recognition.
\newblock In \emph{Proceedings of the IEEE conference on computer vision and pattern recognition}, pages 770--778, 2016.

\bibitem[Hu et~al.(2024)Hu, Cheng, Lu, Cao, Wei, Liu, and Li]{hu2024lf}
Y.~Hu, Y.~Cheng, A.~Lu, Z.~Cao, D.~Wei, J.~Liu, and Z.~Li.
\newblock Lf-vit: Reducing spatial redundancy in vision transformer for efficient image recognition.
\newblock In \emph{Proceedings of the AAAI Conference on Artificial Intelligence}, volume~38, pages 2274--2284, 2024.

\bibitem[Ignatov et~al.(2019)Ignatov, Timofte, Kulik, Yang, Wang, Baum, Wu, Xu, and Van~Gool]{ignatov2019ai}
A.~Ignatov, R.~Timofte, A.~Kulik, S.~Yang, K.~Wang, F.~Baum, M.~Wu, L.~Xu, and L.~Van~Gool.
\newblock Ai benchmark: All about deep learning on smartphones in 2019.
\newblock In \emph{2019 IEEE/CVF International Conference on Computer Vision Workshop (ICCVW)}, pages 3617--3635. IEEE, 2019.

\bibitem[Jiang et~al.(2021)Jiang, Hou, Yuan, Zhou, Shi, Jin, Wang, and Feng]{NEURIPS2021_9a49a25d}
Z.-H. Jiang, Q.~Hou, L.~Yuan, D.~Zhou, Y.~Shi, X.~Jin, A.~Wang, and J.~Feng.
\newblock All tokens matter: Token labeling for training better vision transformers.
\newblock In M.~Ranzato, A.~Beygelzimer, Y.~Dauphin, P.~Liang, and J.~W. Vaughan, editors, \emph{Advances in Neural Information Processing Systems}, volume~34, pages 18590--18602, 2021.

\bibitem[Khanam and Hussain(2024)]{khanam2024yolov11}
R.~Khanam and M.~Hussain.
\newblock Yolov11: An overview of the key architectural enhancements.
\newblock \emph{arXiv preprint arXiv:2410.17725}, 2024.

\bibitem[Kim et~al.(2022)Kim, Kim, Oh, Yun, Song, Jeong, Ha, and Song]{kim2022dataset}
J.-H. Kim, J.~Kim, S.~J. Oh, S.~Yun, H.~Song, J.~Jeong, J.-W. Ha, and H.~O. Song.
\newblock Dataset condensation via efficient synthetic-data parameterization.
\newblock In \emph{International Conference on Machine Learning}, pages 11102--11118. PMLR, 2022.

\bibitem[Kirillov et~al.(2023)Kirillov, Mintun, Ravi, Mao, Rolland, Gustafson, Xiao, Whitehead, Berg, Lo, et~al.]{kirillov2023segment}
A.~Kirillov, E.~Mintun, N.~Ravi, H.~Mao, C.~Rolland, L.~Gustafson, T.~Xiao, S.~Whitehead, A.~C. Berg, W.-Y. Lo, et~al.
\newblock Segment anything.
\newblock In \emph{Proceedings of the IEEE/CVF International Conference on Computer Vision}, pages 4015--4026, 2023.

\bibitem[Krizhevsky et~al.(2009)Krizhevsky, Hinton, et~al.]{krizhevsky2009learning}
A.~Krizhevsky, G.~Hinton, et~al.
\newblock Learning multiple layers of features from tiny images.
\newblock 2009.

\bibitem[Le and Yang(2015)]{le2015tiny}
Y.~Le and X.~Yang.
\newblock Tiny imagenet visual recognition challenge.
\newblock \emph{CS 231N}, 7\penalty0 (7):\penalty0 3, 2015.

\bibitem[Lee et~al.(2022)Lee, Chun, Jung, Yun, and Yoon]{lee2022dataset}
S.~Lee, S.~Chun, S.~Jung, S.~Yun, and S.~Yoon.
\newblock Dataset condensation with contrastive signals.
\newblock In \emph{International Conference on Machine Learning}, pages 12352--12364. PMLR, 2022.

\bibitem[Lin et~al.(2014)Lin, Maire, Belongie, Hays, Perona, Ramanan, Doll{\'a}r, and Zitnick]{lin2014microsoft}
T.-Y. Lin, M.~Maire, S.~Belongie, J.~Hays, P.~Perona, D.~Ramanan, P.~Doll{\'a}r, and C.~L. Zitnick.
\newblock Microsoft coco: Common objects in context.
\newblock In \emph{Computer Vision--ECCV 2014: 13th European Conference, Zurich, Switzerland, September 6-12, 2014, Proceedings, Part V 13}, pages 740--755. Springer, 2014.

\bibitem[Liu et~al.(2022)Liu, Wang, Yang, Ye, and Wang]{liu2022dataset}
S.~Liu, K.~Wang, X.~Yang, J.~Ye, and X.~Wang.
\newblock Dataset distillation via factorization.
\newblock \emph{Advances in neural information processing systems}, 35:\penalty0 1100--1113, 2022.

\bibitem[Liu et~al.(2021)Liu, Lin, Cao, Hu, Wei, Zhang, Lin, and Guo]{liu2021swin}
Z.~Liu, Y.~Lin, Y.~Cao, H.~Hu, Y.~Wei, Z.~Zhang, S.~Lin, and B.~Guo.
\newblock Swin transformer: Hierarchical vision transformer using shifted windows.
\newblock In \emph{Proceedings of the IEEE/CVF international conference on computer vision}, pages 10012--10022, 2021.

\bibitem[Loo et~al.()Loo, Maalouf, Hasani, Lechner, Amini, and Rus]{loolarge}
N.~Loo, A.~Maalouf, R.~Hasani, M.~Lechner, A.~Amini, and D.~Rus.
\newblock Large scale dataset distillation with domain shift.
\newblock In \emph{Forty-first International Conference on Machine Learning}.

\bibitem[Lu et~al.(2023)Lu, Chen, Zhang, Gu, Zhang, Zhang, Yang, Xuan, Wang, and You]{lu2023can}
Y.~Lu, X.~Chen, Y.~Zhang, J.~Gu, T.~Zhang, Y.~Zhang, X.~Yang, Q.~Xuan, K.~Wang, and Y.~You.
\newblock Can pre-trained models assist in dataset distillation?
\newblock \emph{arXiv preprint arXiv:2310.03295}, 2023.

\bibitem[Meding et~al.(2021)Meding, Buschoff, Geirhos, and Wichmann]{meding2021trivial}
K.~Meding, L.~M.~S. Buschoff, R.~Geirhos, and F.~A. Wichmann.
\newblock Trivial or impossible--dichotomous data difficulty masks model differences (on imagenet and beyond).
\newblock \emph{arXiv preprint arXiv:2110.05922}, 2021.

\bibitem[Nguyen et~al.(2021)Nguyen, Novak, Xiao, and Lee]{nguyen2021dataset}
T.~Nguyen, R.~Novak, L.~Xiao, and J.~Lee.
\newblock Dataset distillation with infinitely wide convolutional networks.
\newblock \emph{Advances in Neural Information Processing Systems}, 34:\penalty0 5186--5198, 2021.

\bibitem[Nilsback and Zisserman(2008)]{nilsback2008automated}
M.-E. Nilsback and A.~Zisserman.
\newblock Automated flower classification over a large number of classes.
\newblock In \emph{2008 Sixth Indian conference on computer vision, graphics \& image processing}, pages 722--729. IEEE, 2008.

\bibitem[Paul et~al.(2021)Paul, Ganguli, and Dziugaite]{paul2021deep}
M.~Paul, S.~Ganguli, and G.~K. Dziugaite.
\newblock Deep learning on a data diet: Finding important examples early in training.
\newblock \emph{Advances in Neural Information Processing Systems}, 34:\penalty0 20596--20607, 2021.

\bibitem[Peebles and Xie(2023)]{peebles2023scalable}
W.~Peebles and S.~Xie.
\newblock Scalable diffusion models with transformers.
\newblock In \emph{Proceedings of the IEEE/CVF International Conference on Computer Vision}, pages 4195--4205, 2023.

\bibitem[Radford et~al.(2021)Radford, Kim, Hallacy, Ramesh, Goh, Agarwal, Sastry, Askell, Mishkin, Clark, et~al.]{radford2021learning}
A.~Radford, J.~W. Kim, C.~Hallacy, A.~Ramesh, G.~Goh, S.~Agarwal, G.~Sastry, A.~Askell, P.~Mishkin, J.~Clark, et~al.
\newblock Learning transferable visual models from natural language supervision.
\newblock In \emph{International conference on machine learning}, pages 8748--8763. PMLR, 2021.

\bibitem[Rao et~al.(2021)Rao, Zhao, Liu, Lu, Zhou, and Hsieh]{rao2021dynamicvit}
Y.~Rao, W.~Zhao, B.~Liu, J.~Lu, J.~Zhou, and C.-J. Hsieh.
\newblock Dynamicvit: Efficient vision transformers with dynamic token sparsification.
\newblock \emph{Advances in neural information processing systems}, 34:\penalty0 13937--13949, 2021.

\bibitem[Ridnik et~al.(2021)Ridnik, Ben-Baruch, Noy, and Zelnik-Manor]{ridnik2021imagenet}
T.~Ridnik, E.~Ben-Baruch, A.~Noy, and L.~Zelnik-Manor.
\newblock Imagenet-21k pretraining for the masses.
\newblock \emph{arXiv preprint arXiv:2104.10972}, 2021.

\bibitem[Sandler et~al.(2018)Sandler, Howard, Zhu, Zhmoginov, and Chen]{sandler2018mobilenetv2}
M.~Sandler, A.~Howard, M.~Zhu, A.~Zhmoginov, and L.-C. Chen.
\newblock Mobilenetv2: Inverted residuals and linear bottlenecks.
\newblock In \emph{Proceedings of the IEEE conference on computer vision and pattern recognition}, pages 4510--4520, 2018.

\bibitem[Shao et~al.(2023)Shao, Yin, Zhou, Zhang, and Shen]{shao2023generalized}
S.~Shao, Z.~Yin, M.~Zhou, X.~Zhang, and Z.~Shen.
\newblock Generalized large-scale data condensation via various backbone and statistical matching.
\newblock \emph{arXiv preprint arXiv:2311.17950}, 2023.

\bibitem[Shen and Xing(2022)]{shen2022fast}
Z.~Shen and E.~Xing.
\newblock A fast knowledge distillation framework for visual recognition.
\newblock In \emph{European conference on computer vision}, pages 673--690, 2022.

\bibitem[Su et~al.(2024)Su, Hou, Li, Togo, Song, Ogawa, and Haseyama]{su2024generative}
D.~Su, J.~Hou, G.~Li, R.~Togo, R.~Song, T.~Ogawa, and M.~Haseyama.
\newblock Generative dataset distillation based on diffusion model.
\newblock \emph{arXiv preprint arXiv:2408.08610}, 2024.

\bibitem[Sun et~al.(2024)Sun, Shi, Yu, and Lin]{sun2024diversity}
P.~Sun, B.~Shi, D.~Yu, and T.~Lin.
\newblock On the diversity and realism of distilled dataset: An efficient dataset distillation paradigm.
\newblock In \emph{Proceedings of the IEEE/CVF Conference on Computer Vision and Pattern Recognition}, 2024.

\bibitem[Szegedy et~al.(2015)Szegedy, Liu, Jia, Sermanet, Reed, Anguelov, Erhan, Vanhoucke, and Rabinovich]{szegedy2015going}
C.~Szegedy, W.~Liu, Y.~Jia, P.~Sermanet, S.~Reed, D.~Anguelov, D.~Erhan, V.~Vanhoucke, and A.~Rabinovich.
\newblock Going deeper with convolutions.
\newblock In \emph{Proceedings of the IEEE CVPR}, pages 1--9, 2015.

\bibitem[Tan et~al.(2024)Tan, Wu, Du, Chen, Wang, Wang, and Qi]{tan2024data}
H.~Tan, S.~Wu, F.~Du, Y.~Chen, Z.~Wang, F.~Wang, and X.~Qi.
\newblock Data pruning via moving-one-sample-out.
\newblock \emph{Advances in Neural Information Processing Systems}, 36, 2024.

\bibitem[Tan and Le(2019)]{tan2019efficientnet}
M.~Tan and Q.~Le.
\newblock Efficientnet: Rethinking model scaling for convolutional neural networks.
\newblock In \emph{International conference on machine learning}, pages 6105--6114. PMLR, 2019.

\bibitem[Toneva et~al.(2018)Toneva, Sordoni, Combes, Trischler, Bengio, and Gordon]{toneva2018empirical}
M.~Toneva, A.~Sordoni, R.~T.~d. Combes, A.~Trischler, Y.~Bengio, and G.~J. Gordon.
\newblock An empirical study of example forgetting during deep neural network learning.
\newblock \emph{arXiv preprint arXiv:1812.05159}, 2018.

\bibitem[Touvron et~al.(2021)Touvron, Cord, Douze, Massa, Sablayrolles, and J{\'e}gou]{touvron2021training}
H.~Touvron, M.~Cord, M.~Douze, F.~Massa, A.~Sablayrolles, and H.~J{\'e}gou.
\newblock Training data-efficient image transformers \& distillation through attention.
\newblock In \emph{International conference on machine learning}, pages 10347--10357. PMLR, 2021.

\bibitem[Vaswani et~al.(2017)Vaswani, Shazeer, Parmar, Uszkoreit, Jones, Gomez, Kaiser, and Polosukhin]{vaswani2017attention}
A.~Vaswani, N.~Shazeer, N.~Parmar, J.~Uszkoreit, L.~Jones, A.~N. Gomez, {\L}.~Kaiser, and I.~Polosukhin.
\newblock Attention is all you need.
\newblock \emph{Advances in neural information processing systems}, 30, 2017.

\bibitem[Wang et~al.(2022)Wang, Zhao, Peng, Zhu, Yang, Wang, Huang, Bilen, Wang, and You]{wang2022cafe}
K.~Wang, B.~Zhao, X.~Peng, Z.~Zhu, S.~Yang, S.~Wang, G.~Huang, H.~Bilen, X.~Wang, and Y.~You.
\newblock Cafe: Learning to condense dataset by aligning features.
\newblock In \emph{2022 IEEE/CVF Conference on Computer Vision and Pattern Recognition (CVPR)}, pages 12186--12195, 2022.
\newblock \doi{10.1109/CVPR52688.2022.01188}.

\bibitem[Wang et~al.(2018)Wang, Zhu, Torralba, and Efros]{wang2018dataset}
T.~Wang, J.-Y. Zhu, A.~Torralba, and A.~A. Efros.
\newblock Dataset distillation.
\newblock \emph{arXiv preprint arXiv:1811.10959}, 2018.

\bibitem[Welling(2009)]{welling2009herding}
M.~Welling.
\newblock Herding dynamical weights to learn.
\newblock In \emph{Proceedings of the 26th annual international conference on machine learning}, pages 1121--1128, 2009.

\bibitem[Yang et~al.(2024)Yang, Zhu, Deng, and Russakovsky]{yang2024dataset}
W.~Yang, Y.~Zhu, Z.~Deng, and O.~Russakovsky.
\newblock What is dataset distillation learning?
\newblock \emph{arXiv preprint arXiv:2406.04284}, 2024.

\bibitem[Yin et~al.(2019)Yin, Kannan, and Bartlett]{yin2019rademacher}
D.~Yin, R.~Kannan, and P.~Bartlett.
\newblock Rademacher complexity for adversarially robust generalization.
\newblock In \emph{International conference on machine learning}, pages 7085--7094. PMLR, 2019.

\bibitem[Yin and Shen(2023)]{yin2023dataset}
Z.~Yin and Z.~Shen.
\newblock Dataset distillation in large data era.
\newblock \emph{arXiv e-prints}, pages arXiv--2311, 2023.

\bibitem[Yin et~al.(2024)Yin, Xing, and Shen]{yin2024squeeze}
Z.~Yin, E.~Xing, and Z.~Shen.
\newblock Squeeze, recover and relabel: Dataset condensation at imagenet scale from a new perspective.
\newblock \emph{Advances in Neural Information Processing Systems}, 36, 2024.

\bibitem[Yun et~al.(2021)Yun, Oh, Heo, Han, Choe, and Chun]{yun2021re}
S.~Yun, S.~J. Oh, B.~Heo, D.~Han, J.~Choe, and S.~Chun.
\newblock Re-labeling imagenet: from single to multi-labels, from global to localized labels.
\newblock In \emph{Proceedings of the IEEE/CVF Conference on Computer Vision and Pattern Recognition}, pages 2340--2350, 2021.

\bibitem[Zhang et~al.(2023)Zhang, Zhang, Lei, Mukherjee, Pan, Zhao, Ding, Li, and Xu]{zhang2023accelerating}
L.~Zhang, J.~Zhang, B.~Lei, S.~Mukherjee, X.~Pan, B.~Zhao, C.~Ding, Y.~Li, and D.~Xu.
\newblock Accelerating dataset distillation via model augmentation.
\newblock In \emph{Proceedings of the IEEE/CVF Conference on Computer Vision and Pattern Recognition}, pages 11950--11959, 2023.

\bibitem[Zhao and Bilen(2021)]{zhao2021dataset}
B.~Zhao and H.~Bilen.
\newblock Dataset condensation with differentiable siamese augmentation.
\newblock In \emph{International Conference on Machine Learning}, pages 12674--12685. PMLR, 2021.

\bibitem[Zhao and Bilen(2022)]{zhao2022synthesizing}
B.~Zhao and H.~Bilen.
\newblock Synthesizing informative training samples with gan.
\newblock \emph{arXiv preprint arXiv:2204.07513}, 2022.

\bibitem[Zhao and Bilen(2023)]{zhao2023dataset}
B.~Zhao and H.~Bilen.
\newblock Dataset condensation with distribution matching.
\newblock In \emph{Proceedings of the IEEE/CVF Winter Conference on Applications of Computer Vision}, pages 6514--6523, 2023.

\bibitem[Zhao et~al.(2020)Zhao, Mopuri, and Bilen]{zhao2020dataset}
B.~Zhao, K.~R. Mopuri, and H.~Bilen.
\newblock Dataset condensation with gradient matching.
\newblock \emph{arXiv preprint arXiv:2006.05929}, 2020.

\bibitem[Zhou et~al.(2024)Zhou, Yin, Shao, and Shen]{SCDD}
M.~Zhou, Z.~Yin, S.~Shao, and Z.~Shen.
\newblock Self-supervised dataset distillation: A good compression is all you need.
\newblock \emph{arXiv preprint arXiv:2404.07976}, 2024.

\bibitem[Zhou et~al.(2022)Zhou, Nezhadarya, and Ba]{zhou2022dataset}
Y.~Zhou, E.~Nezhadarya, and J.~Ba.
\newblock Dataset distillation using neural feature regression.
\newblock \emph{arXiv 2206.00719}, 2022.

\end{thebibliography}
